\documentclass[12pt]{article}
\usepackage{graphicx}
\usepackage{amsmath,amssymb} % define this before the line numbering.
\usepackage{color}
\usepackage{verbatim}
\usepackage{csvsimple}

\newcommand{\Iter}[1]{^{({#1})}}

\begin{document}

\title{Fast and easy blind deblurring using an inverse filter and PROBE}

\author{Naftali Zon $~~~$ Rana Hanocka $~~~$ Nahum Kiryati\\[0.2cm]
School of Electrical Engineering\\
Tel Aviv University\\
Tel Aviv 69978, Israel}

\date{}

\maketitle

\begin{abstract}
PROBE  (\textbf{P}rogressive \textbf{R}emoval \textbf{o}f \textbf{B}lur R\textbf{e}sidual) is a recursive framework for blind deblurring. Using the elementary modified inverse filter at its core, PROBE's experimental performance meets or exceeds the state of the art, both visually and quantitatively. Remarkably, PROBE lends itself to analysis that reveals its convergence properties.
PROBE is motivated by recent ideas on progressive blind deblurring,
but breaks away from previous research by its simplicity, speed, performance and potential for analysis.
PROBE is neither a functional minimization approach, nor an open-loop sequential
method (blur kernel estimation followed by non-blind deblurring). PROBE is a feedback scheme, deriving its unique strength from the closed-loop architecture rather than from the accuracy of its algorithmic components.
% Thus, PROBE requires less tuning than previous methods.
%\keywords{blind deblurring, pseudo-inverse filter, residual blur elimination, PROBE framework}
\end{abstract}

\section{Introduction}
Despite decades of research, image deblurring remains a highly challenging problem.
Defocus blur and motion blur are both modelled as a linear process
\begin{equation}\label{eq:linear}
g = u*h + n,
\end{equation}
where the blurred image $g$ is the convolution of an unknown latent image $u$ with a blur kernel $h$ plus noise $n$.
The deblurring problem is referred to as non-blind or blind, depending on whether $h$ is known or unknown respectively.

Non-blind deblurring, {\em i.e.} the recovery of $u$ given $g$ and $h$,  can be formulated as a functional minimization problem
\begin{equation}\label{eq:nonblind}
\hat{u} = \min_{u} \frac{\gamma}{2} ||h*u - g||_{2}^{2} + Q(u).
\end{equation}
The data term $||h*u - g||_{2}^{2}$ reflects Gaussian noise, and the regularization term $Q(u)$, needed since the
problem is ill-posed, represents {\em a-priori} information about image structure.
This formulation is equivalent to a maximum a-posteriori (MAP) statistical estimation problem, via the negative log-likelihood.

Mainstream blind deblurring schemes fall into two major categories: {\em sequential} approaches first estimate the blur kernel then employ non-blind deblurring; {\em parallel} approaches simultaneously estimate the latent image and the blur kernel.  Sequential approaches which utilize functional minimization for the non-blind deblurring phase, do so as in equation~(\ref{eq:nonblind}).
Parallel approaches that follow the energy minimization paradigm aim to minimize a joint functional
\begin{equation} \label{eq:MAPuh}
(\hat{u},\hat{h})_{\textrm{MAP}} = \min_{u,h} \frac{\gamma}{2} ||h*u - g||_{2}^{2} + Q(u) + R(h),
\end{equation}
where the blur regularizer $R(h)$ represents {\em a-priori} information about the point spread function (PSF).
This is equivalent to maximizing the posterior
\begin{equation}
(\hat{u},\hat{h})_{\textrm{MAP}} = \max_{u,h} p(u,h|g) = \max_{u,h} p(g|u,h)p(h)p(u),
\end{equation}
and is referred to as (parallel) MAP$_{u,h}$.
The variational problem~(\ref{eq:MAPuh}) is solved via alternate minimization (AM);
the update equations for $u$ and $h$ are
\begin{equation} \label{eq:updateUH}
\begin{aligned}
u^{l+1} \leftarrow \min_{u} \frac{\gamma}{2} ||h^{l} * u - g||_{2}^{2} + Q(u) \\
h^{l + 1} \leftarrow \min_{h} \frac{\gamma}{2} ||h * u^{l} - g||_{2}^{2} + R(h)
\end{aligned}
\end{equation}
where the superscript $l$ is the iteration number.

True parallel MAP$_{u,h}$ approaches, described in~\cite{ChanWong,YouKaveh}, are mathematically elegant, but have been observed~\cite{bar2005,fergus2006,levin2009,perrone2014} to fail in practice. Levin {\em et al}~\cite{levin2009} demonstrated that simultaneous estimation of the latent image and the blur kernel, using the parallel MAP$_{u,h}$  approach with common image priors $Q(u)$, actually favors the trivial solution $\hat{u} = g$ and $\hat{h} = \delta$. This means that the observed blur in $g$ is assumed to be due to blur in the latent image $u$, rather than due to the blur process $h$. Such assumption obviously results in a blurred solution $\hat{u}$. Therefore, the true MAP$_{u,h}$ minimizers $(\hat{u},\hat{h})_{\textrm{MAP}}$ are {\em not} the desired solution.

Works which claim to follow the true MAP$_{u,h}$ paradigm, tacitly avoid the undesired true minimizers using ad-hoc steps to minimize a non-equivalent functional, that we refer to as parallel ad-hoc MAP$_{u,h}$. Parallel ad-hoc MAP$_{u,h}$ approaches usually increase the likelihood weight ($\gamma$ in equation~(\ref{eq:MAPuh})) during minimization and apply the blur kernel constraints via sequential-projection (see~\cite{perrone2014}). For instance, Perrone and Favaro~\cite{perrone2014} showed that in practice~\cite{ChanWong} used ad-hoc steps, which do not actually minimize the true MAP$_{u,h}$. Levin {\em et al}~\cite{levin2009} demonstrated the same for~\cite{shan2008}. While ad-hoc steps push the estimators away from the true MAP$_{u,h}$ minimizers~\cite{perrone2014,levin2009}, this often results in the desired solution~\cite{levin2009}.

While parallel ad-hoc MAP$_{u,h}$ is a viable alternative to true MAP$_{u,h}$, recent works have had better success utilizing the sequential approach~\cite{kotera2013,krishnan2011,komodakis2012}. Levin {\em et al}~\cite{levin2009} showed that the sequential approach is more stable and a lower dimensional problem.
We name the sequential class of algorithms which use the MAP paradigm for kernel estimation as ad-hoc MAP$_{u_c,h}$, where $u_c \neq u$. $u_c$ is typically a lower dimensional representation of $u$, such as a {\em cartoon-image}, discarded after AM.
Ad-hoc MAP$_{u_c,h}$ is always followed by a separate non-blind deblurring phase (sequential approach). MAP$_{u_c,h}$ usually uses the same ad-hoc steps as described in the parallel ad-hoc MAP$_{u,h}$ approaches, but additionally uses unnatural, exceedingly sparse priors~\cite{SCho_deblur_2009,xu2010,kotera2013} to obtain the cartoon image $u_c$.
Levin {\em et al}~\cite{levin2009} showed that a cartoon-like image results from increasing the likelihood weight.

\begin{figure}[t]
\centering
\begin{tabular}{ccccc}
\large{\mbox{$g_c$}} & & \large{\mbox{$u_c$}} & & \large{\mbox{$h$}} \\
\raisebox{-.5\height}{\includegraphics[scale=0.2]{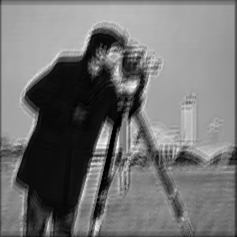}} &
$\textbf{=}$ &
\raisebox{-.5\height}{\includegraphics[scale=0.2]{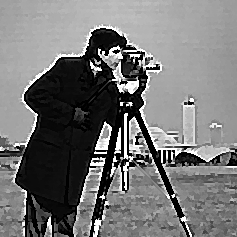}} &
$\textbf{*}$ &
\raisebox{-.5\height}{\includegraphics[scale=2]{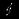}} \\
 & & $L_{0.8} = 7350$ &  &  \\
  & & $ {\color{blue} \Large{\downarrow} } $ &  &  \\
\large{\mbox{$g$}} & & \large{\mbox{$u$}} & & \large{\mbox{$h$}} \\
\raisebox{-.5\height}{\includegraphics[scale=0.2]{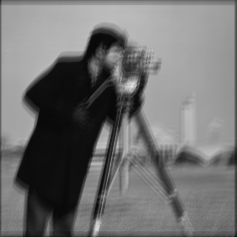}} &
$\textbf{=}$ &
\raisebox{-.5\height}{\includegraphics[scale=0.2]{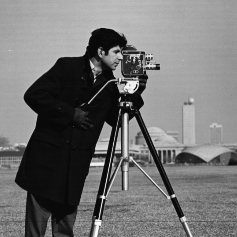}} &
$\textbf{*}$ &
\raisebox{-.5\height}{\includegraphics[scale=2]{h.png}} \\
 & & $L_{0.8} = 4866$ &  &
\end{tabular}
\caption{Ad-hoc MAP$_{u_c,h}$ for kernel estimation. Cartoon image $u_c$, produced by shock filtering, creates a simplified but equivalent kernel estimation problem ({\em i.e.,} $g = u*h \approx g_c = u_c *h$). The prior energy of $u_c$ is greater than $u$, which helps push MAP$_{u_c,h}$ away from the trivial solution $h=\delta$ and towards the desired blur kernel.}
\label{fig:cartoonFig}
\end{figure}

The cartoon image emphasizes salient edges and suppresses weak details in flat regions~\cite{elad2015}. It is useful for accurate kernel estimation, conceivably since $u_c$ directs the kernel estimator to stronger step-edges, diverting away from weak edges possibly related to noise. Thus, the cartoon image functions as a regularizer.
From another perspective, ad-hoc MAP$_{u_c,h}$ exploits the ill-posedness of the problem by focusing on an equivalent but easier problem: the outcome of blurring a cartoon image $u_c$ is similar to blurring the true latent image $u$. Formally,
$g_c \approx g $ where $g_c = u_c*h $ and $g = u*h$. This has been observed in~\cite{komodakis2012} for discrete images.
The cartooning  effect can simply be achieved via shock filtering~\cite{Osher1990}. This can be used to steer MAP$_{u_c,h}$ towards the
desired blur kernel, as shown in Fig.~\ref{fig:cartoonFig}.

The empirical failure of the parallel MAP$_{u,h}$ approach, and its explanation by Levin {\em et al}~\cite{levin2009}, have driven the blind deblurring field towards the sequential deblurring framework, where blur kernel estimation is followed by non-blind deblurring. However, the sequential framework is limited, since any inaccuracy in either PSF estimation or non-blind deblurring leads to irreparable deblurring errors. Thus, successful application of the sequential framework requires accurate, sophisticated, complex and computationally expensive PSF estimation and non-blind deblurring algorithms. The proposed PROBE framework is a superior closed-loop alternative to both the parallel and sequential approaches.

\section{PROBE Framework}

In culinary arts, it is said that a dish can only be as good as its ingredients. In blind deblurring recipes, the ingredients are blur-kernel estimation and non-blind deblurring. Sequential approaches apply these ingredients consecutively; parallel
MAP$_{u,h}$ approaches apply them iteratively. Typically, researchers tend to use the best algorithmic ingredients at their access.
Recently, Hanocka and Kiryati~\cite{hanocka2015} presented a successful blind deblurring algorithm, based on sophisticated and computationally expensive ingredients. For blur-kernel estimation they extracted an ad-hoc MAP$_{u_c,h}$ algorithm from the implementation of Kotera {\em et al}~\cite{kotera2013}. For non-blind deblurring they adopted the non-blind version of Bar {\em et al}~\cite{bar2005}, based on the $\gamma$-convergence approximation of Mumford-Shah regularization.

We claim that the successful result of~\cite{hanocka2015} is primarily due to its special architecture, rather than due to the sophisticated algorithmic ingredients used. We specify this architecture and call it PROBE, shorthand for
\textbf{P}rogressive \textbf{R}emoval \textbf{o}f \textbf{B}lur R\textbf{e}sidual. We show that~\cite{hanocka2015} is an overly complicated and needlessly expensive version of PROBE. We replace the non-blind deblurring module by a simple modified inverse filter, show that it yields results as good as those of~\cite{hanocka2015} or better, and are similar or superior to the state of the art in general. Furthermore, we show that by simplifying~\cite{hanocka2015}, PROBE not only yields faster processing and better experimental results, but unlike previous works lends itself to analysis. Specifically, we obtain analytic results characterizing the convergence of PROBE. Thus, in contrast to the culinary arts, the PROBE architecture allows the blind deblurring scheme to be better than its non-blind deblurring ingredient.

\begin{figure}[t]
\begin{tabular}{cc}
\includegraphics[width=55mm]{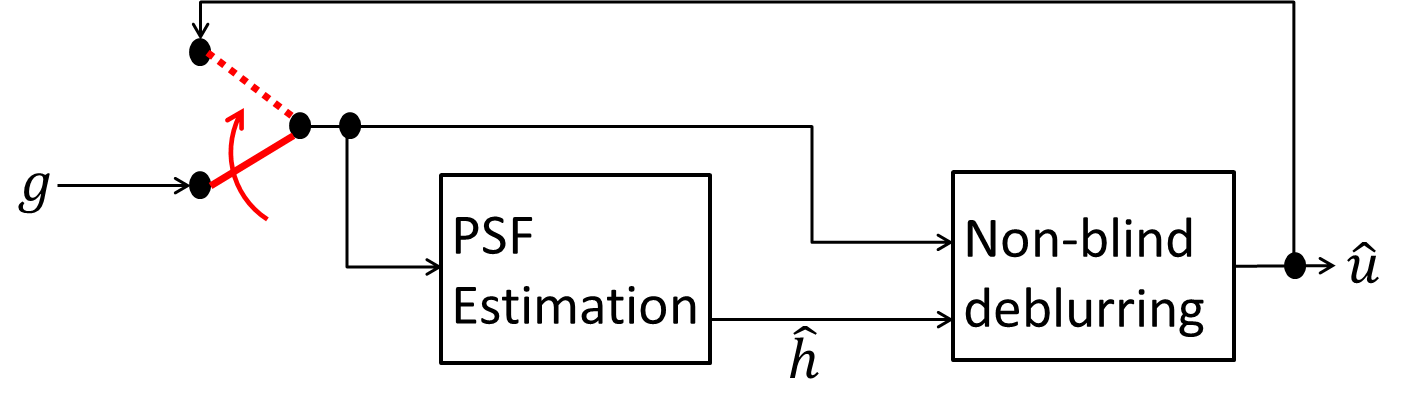} &
\includegraphics[width=55mm]{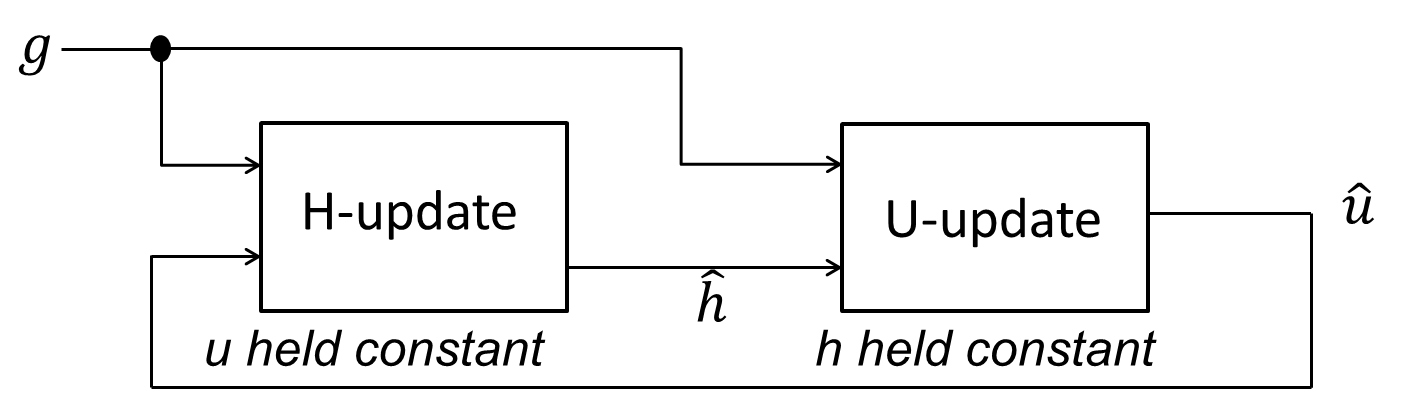} \\
PROBE framework & True MAP$_{u,h}$ \\
\end{tabular}
\caption{Left: PROBE framework adds a feedback loop to the familiar sequential approach to progressively correct for errors. Right: parallel MAP$_{u,h}$ approach which jointly minimizes a unified energy functional.}
\label{fig:probeFramework}
\end{figure}

PROBE is illustrated in Fig.~\ref{fig:probeFramework}(left). Initially, the switch is at the lower position, and the system coincides with the familiar sequential deblurring scheme, consisting of PSF estimation followed by non-blind deblurring. After the first iteration, the switch is thrown to the upper position. Then, the imperfect outcome of the current iteration is fed back as input to the next iteration, {\em i.e.,}
\begin{equation} \label{eq:PROBE_gl}
\begin{aligned}
g^{l} \leftarrow \hat{u}^{l-1} \\
%\textrm{where } \hat{u}^{(l-1)} \overset{!}= g^{(l-1)}*h_{\mathcal(L)}
\end{aligned}
\end{equation}
where superscripts denote iteration numbers.
The PSF estimator identifies the {\em residual blur} remaining in the current input image $g^l$. The non-blind deblurring module removes some of the residual blur, leading to an even better deblurring result $\hat{u}^l$. Recursion continues until a stopping criterion is met. The feedback is the key to PROBE's superior performance: unlike an open-loop system, feedback can potentially correct errors due to imperfect system components.

PROBE is fundamentally different than the parallel MAP$_{u,h}$ approach, illustrated in Fig.~\ref{fig:probeFramework}(right). In PROBE, the input-image for the non-blind deblurring module varies between iterations. In contrast, in parallel MAP$_{u,h}$ the input to the deblurring module is fixed throughout the iterative process. Furthermore, in PROBE the outcome of the PSF estimation module should be the {\em residual} blur, gradually coming close to an impulse function. In parallel MAP$_{u,h}$, the output of the blur estimation module should approach the original blur kernel. Ideally,
\begin{equation}
\begin{aligned}
\lim_{l \rightarrow \infty} \hat{h}_{\mbox{\tiny{MAP}}}^{l}= h \\
\lim_{l \rightarrow \infty} \hat{h}_{\mbox{\tiny{PROBE}}}^{l} = \delta \\
\end{aligned}
\end{equation}

The ability of PROBE's feedback scheme to compensate for component inaccuracies is a key observation. We replace the intricate non-blind deblurring algorithm of~\cite{bar2005} that is used in~\cite{hanocka2015} by the simplest, fastest and crudest practical alternative: the modified inverse filter, given by the transfer function
\begin{equation}\label{eq:pseudo-inverse-generic}
H_{\mbox{\tiny{RI}}} = \frac{H^*}{|H|^2 + C},
\end{equation}
where $H$ is the assumed (current, residual) blur.
$H_{\mbox\tiny{RI}}$ is a smooth approximation of the pseudo-inverse filter. It minimizes the Tikhonov regularized problem
\begin{equation}\label{eq:tikhonov}
\min_u ||u*h - g||^2_2 + C||u||^2_2
\end{equation}
and is also a crude approximation of the Wiener filter with unknown noise statistics, approximated by the constant $C$. Before demonstrating PROBE's superior experimental performance, we mathematically analyze its convergence.

\section{PROBE Convergence} \label{sec:PROBE_analysis}
Using the modified inverse filter as the non-blind restoration module simplifies PROBE substantially. However, effective PSF estimation algorithms are intricate. To facilitate convergence analysis, we model PROBE's blur estimation module by an {\em oracle}, providing the exact blur kernel that relates the input image $g^l$ to the latent image $u$.

PROBE recursively estimates the residual blur kernel and employs non-blind deblurring to reduce the residual blur. The input for each PROBE iteration $l$ is a different blurred image $g^l$ given by
\begin{equation}\label{eq:oracle_problem}
g^l = h_o^l \ast u + n^l,
\end{equation}
where the blur kernel $h_o^l$ provided by the oracle, relates the current image $g^l$ to the latent image $u$. The blur kernel is forwarded to the the modified inverse filter for the non-blind deblurring phase. In the current iteration, the transfer function of the modified inverse filter becomes
\begin{equation}\label{eq:pseudo-inverse}
H_{\mbox{\tiny{RI}}}^l = \frac{(H_o^l)^{*}}{|H_o^l|^{2} + C},
\end{equation}
where $C$ is a regularization parameter and $H_o^l$ is the Fourier transform of $h_o^l$.

Equations~(\ref{eq:oracle_problem}) and~(\ref{eq:pseudo-inverse}) are the basis for convergence analysis. The input $g^{l+1}$ for the coming iteration $l+1$ is a filtered version of $g^l$:
\begin{equation}\label{eq:deblur_pseudo_iters}
g^{l+1} = h_{\mbox{\tiny{RI}}}^l \ast g^l =
h_{\mbox{\tiny{RI}}}^l \ast h_o^l \ast u + h_{\mbox{\tiny{RI}}}^l \ast n^l,
\end{equation}
where $h_{\mbox{\tiny{RI}}}^l$ is the spatial domain form of equation~(\ref{eq:pseudo-inverse}).
In the frequency domain,
\begin{equation}
G^{l+1} =
H_o^l H_{\mbox{\tiny{RI}}}^l U + N^l H_{\mbox{\tiny{RI}}}^l =
\frac{|H_o^l|^2}{|H_o^l|^2 + C}U + N^{l+1}
\end{equation}
where $N^l$ and $U$ are the Fourier transforms of $n^l$ and $u$ respectively.
The blur kernel provided by the oracle in iteration $l+1$ is therefore simply related to the blur kernel the oracle had given in the previous iteration $l$ by
\begin{equation}\label{eq:oracle_recursive}
H_o^{l+1} = \frac{|H_o^l|^2}{|H_o^l|^2 + C}
\end{equation}
at any spatial frequency.

PROBE convergence is indicated by convergence of the blur kernel provided by the oracle, {\em i.e.,}
when $H_o^{l+1} = H_o^l$. Equation~(\ref{eq:oracle_recursive}) implies that $H_o^l$ is strictly positive for any $l>0$.
The final filter $H_o^\infty$ can obtained by solving equation~(\ref{eq:oracle_recursive}) with
$H_o^{l+1} = H_o^l = H_o^\infty$. The resulting equation has three solutions,
$$H_o^\infty = 0 \quad or \quad H_o^\infty \approx C \quad or \quad H_o^\infty \approx 1 -C$$
Note that the large and small roots of $\frac{1\pm\sqrt{1-4C}}{2}$ are approximated by $1 - C$ and $C$, respectively.
Dynamically, for $C>\frac{1}{4}$ the filter converges to zero, while for
$C<\frac{1}{4}$ it can be shown that:
\begin{equation} \label{eq:H_conv}
%\begin{flalign*}
\begin{aligned}
\left\{	\begin{array}{lll}
		 H_o^{l+1} < H_o^{l} &~~& H_o^{0}\in (0,C)\\
		 H_o^{l+1} > H_o^{l} &~~& H_o^{0}\in (C,1-C)\\
		 H_o^{l+1} < H_o^{l} &~~& H_o^{0}\in (1-C,+\infty)
\end{array}
\right.
%\end{flalign*}
\end{aligned}
\end{equation}

The final PROBE blur kernel $H_o^\infty$ at each spatial frequency converges to one of the two stable solutions
$0$ and $1-C$. As seen in equation~(\ref{eq:H_conv}), the third solution $C$ is unstable.
In the case of defocus blur, where the original PSF $H_o^0$ that blurred the latent image can be assumed to be a monotonically decreasing low-pass filter, the final output $H_o^\infty$ of the oracle is an ideal low-pass filter assuming values of either $0$ or $1-C$. For small values of $C$, $H_o^\infty$ approaches $1$ at a wide range of spatial frequencies, and in the limit
\begin{equation} \label{eq:H_delta}
\begin{aligned}
\lim_{C \rightarrow 0} H_o^\infty = 1 , \quad \lim_{C \rightarrow 0} h_o^\infty = \delta.
\end{aligned}
\end{equation}
This is the ideal outcome, since the oracle indicates that all the blur has been removed.
It is not surprising, since for $C \approx 0$, the modified inverse filter resembles the inverse filter.
We will soon see that noise sets a lower limit on useful $C$ values.

\begin{figure}[t]
\center
\begin{tabular}{cc}
Evolution of $H_o^l$ with $l$ & $H_o^\infty$ for decreasing $C$ \\
\includegraphics[width=60mm]{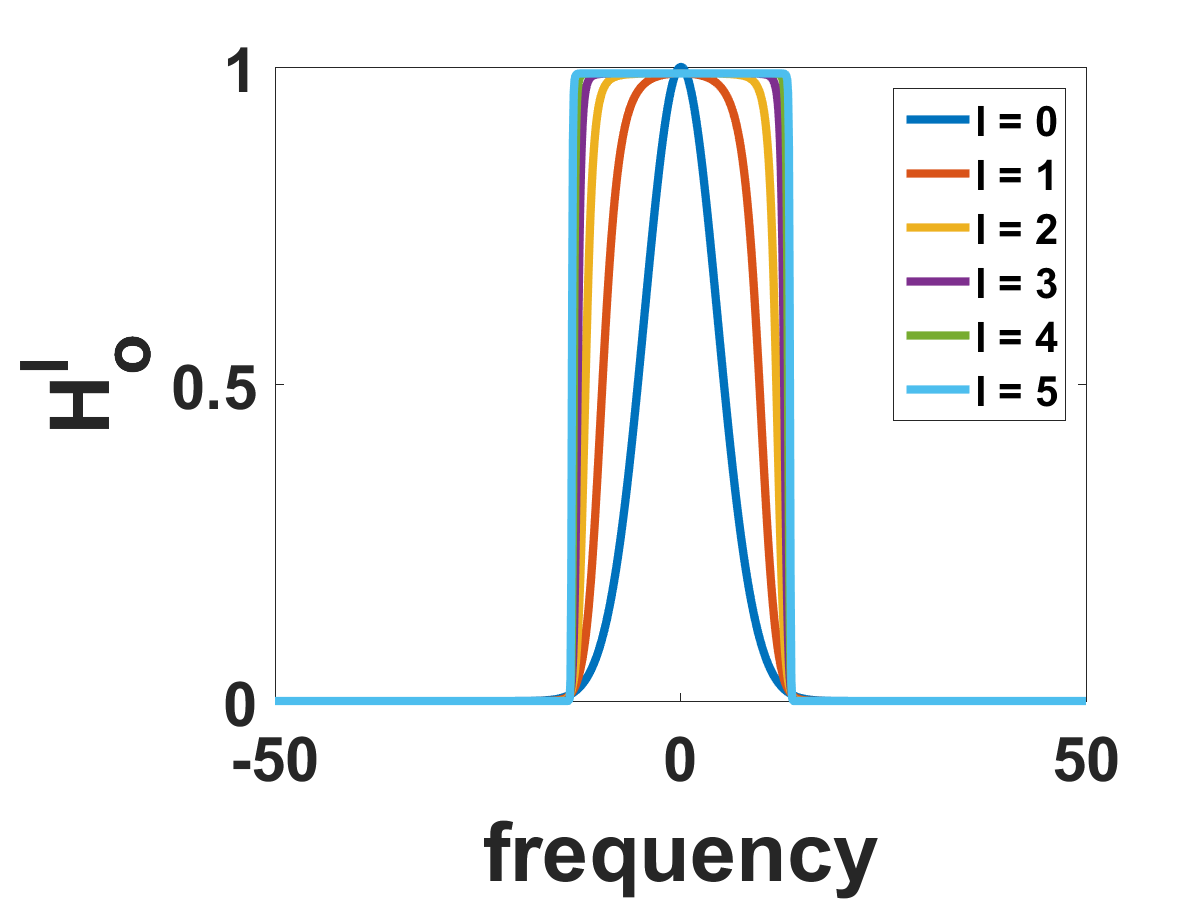} &
\includegraphics[width=60mm]{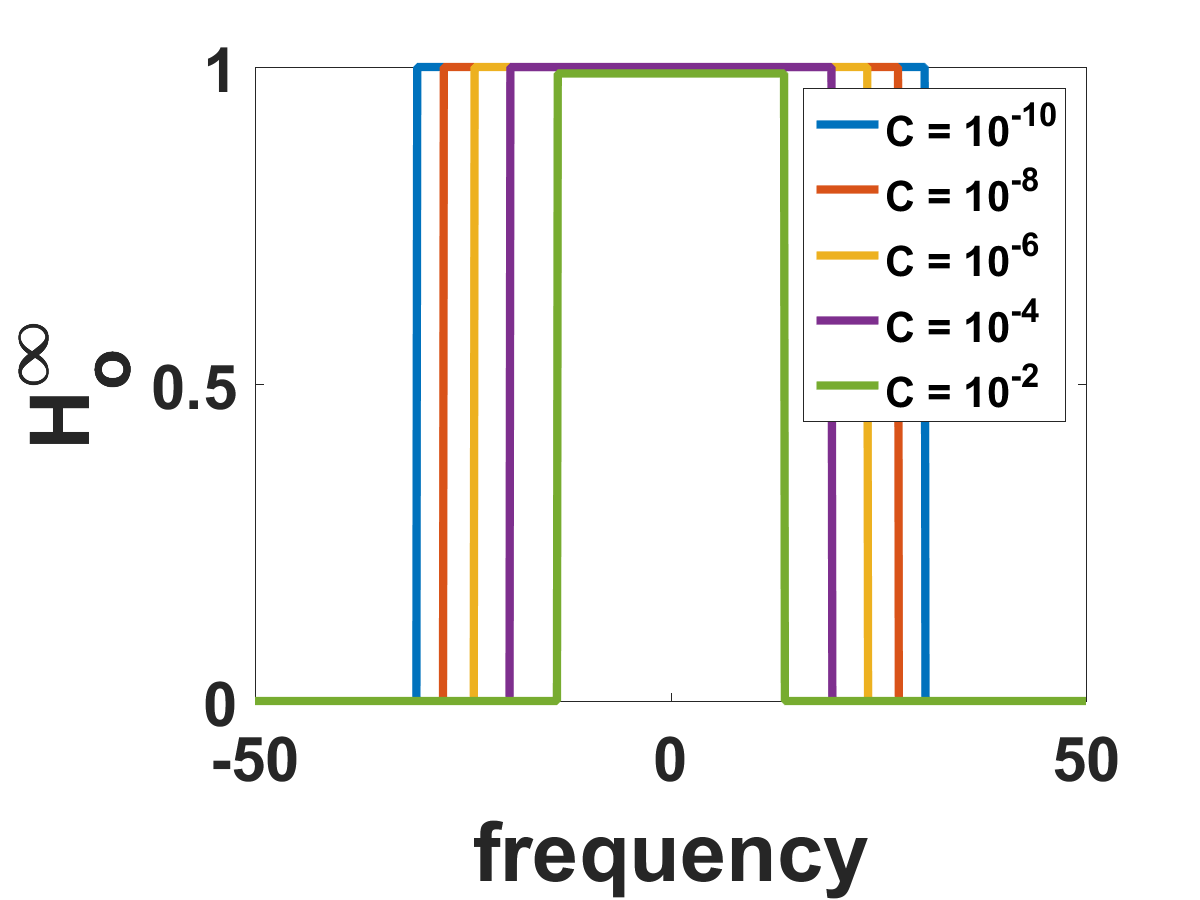} \\
(a) & (b)
\end{tabular}
\caption{(a) Evolution of the oracle-provided PSF $H_o^l$ with PROBE iteration number $l$, starting with a known Gaussian kernel at $l=0$ and converging to an ideal low-pass filter.
(b) $H_o^\infty$ approaches a flat spectrum as $C \rightarrow 0$, corresponding to an impulsive blur kernel $h_o^\infty$ and implying that all blur has been removed.}
\label{fig:oracle_pi}
\end{figure}

The convergence process is visualized for the one-dimensional case in Fig.~\ref{fig:oracle_pi}.
Fig.~\ref{fig:oracle_pi}a shows the evolution of $H_o^l$, corresponding to the PSF estimated by the oracle, over several iterations.
Starting with a Gaussian blur kernel represented by $H_o^0$, and $C = 10^{-2}$, $H_o^l$ converges to an ideal low-pass filter as predicted by the analysis.
Fig.~\ref{fig:oracle_pi}b shows that as $C$ approaches $0$, $H_o^\infty$ approaches a flat spectrum,
meaning the residual blur kernel approaches an impulse function and all blur has been removed.

\begin{figure}[t]
\center
\begin{tabular}{ccc}
 & \large{$H_o^l$ evolution {\em vs} $C$} &  \\
\includegraphics[width=40mm]{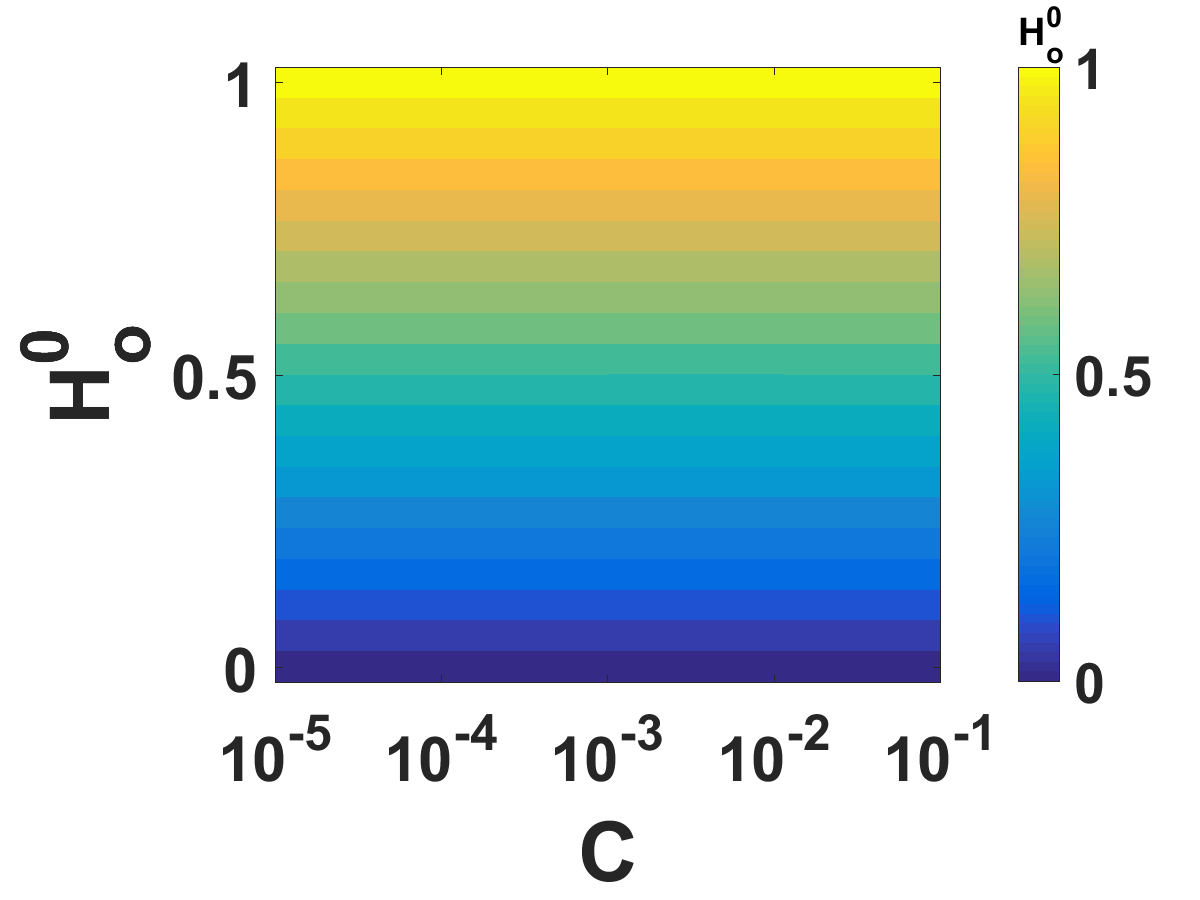} &
\includegraphics[width=40mm]{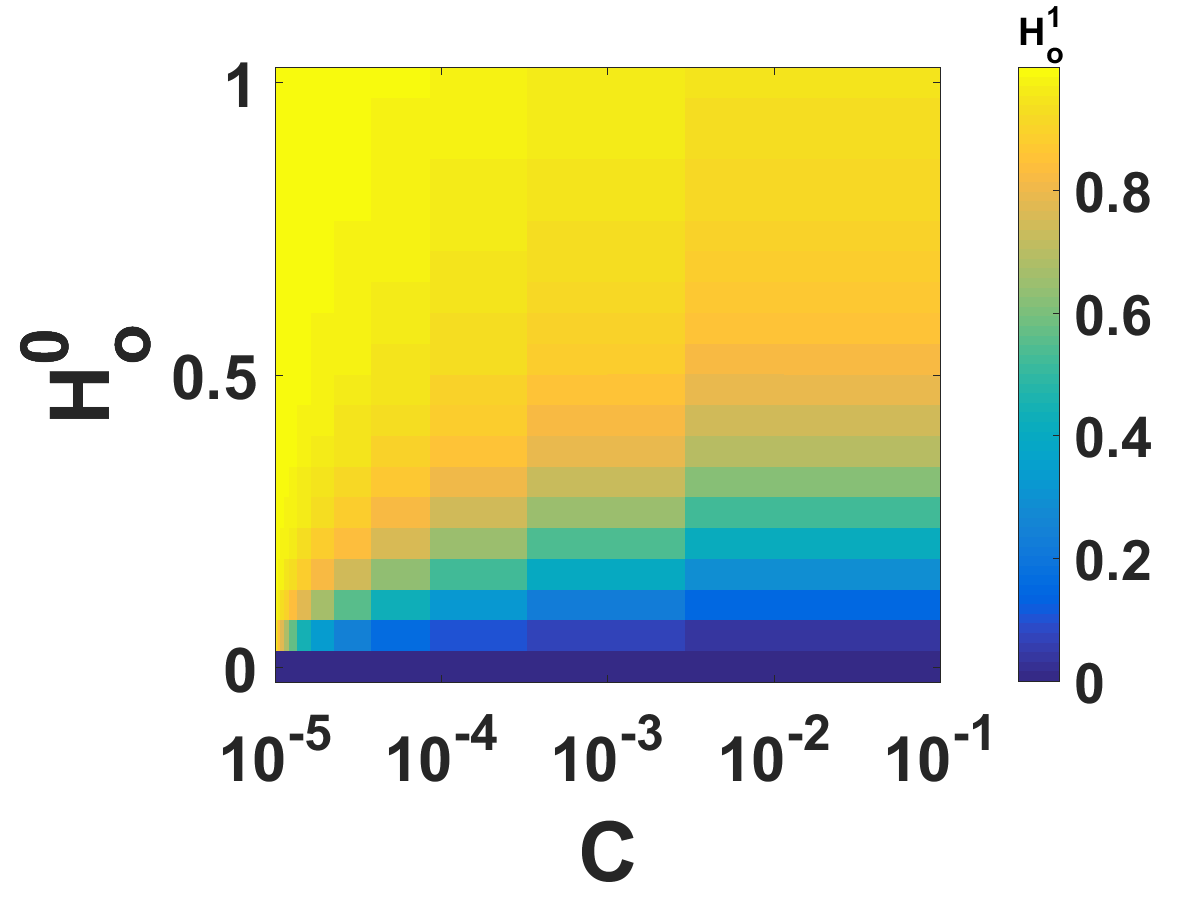} &
\includegraphics[width=40mm]{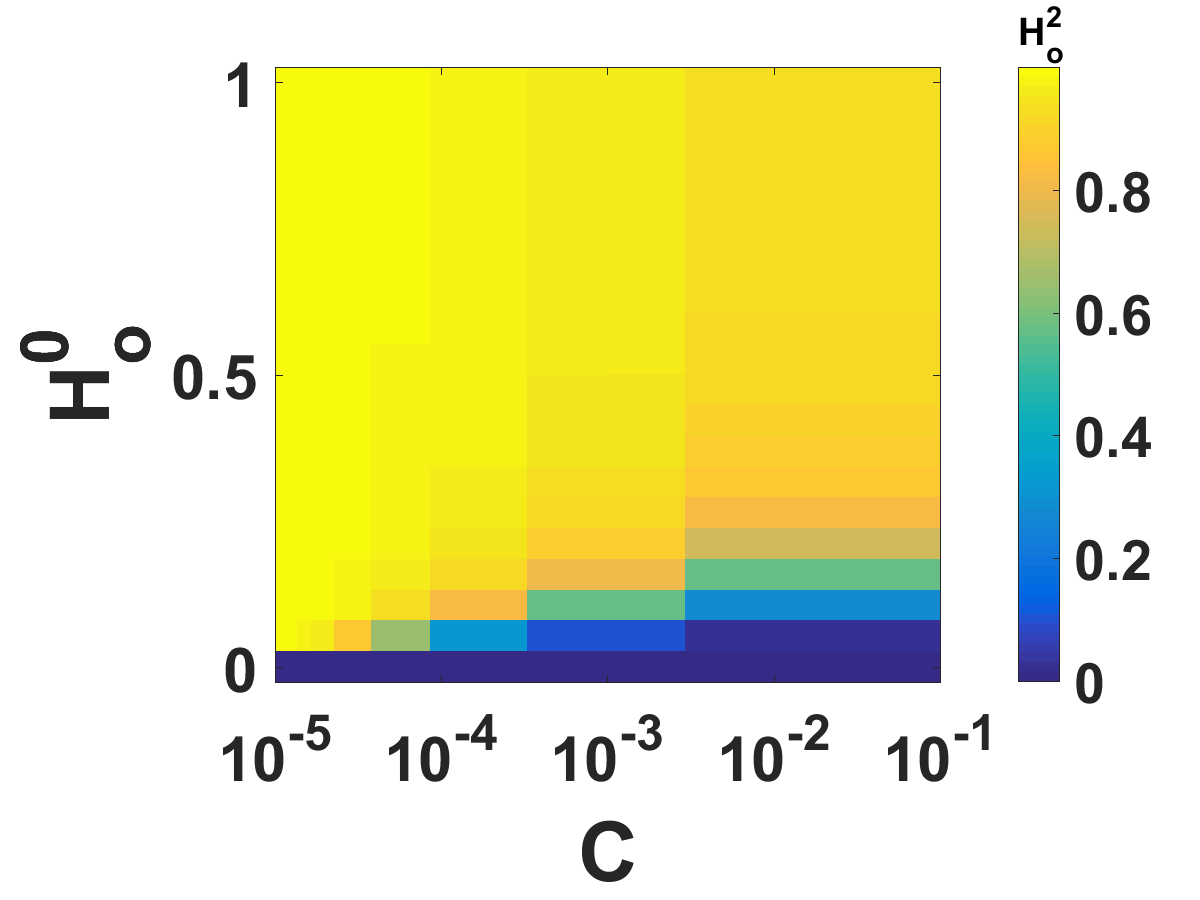} \\
Iteration 0 & Iteration 1 & Iteration 2 \\
\includegraphics[width=40mm]{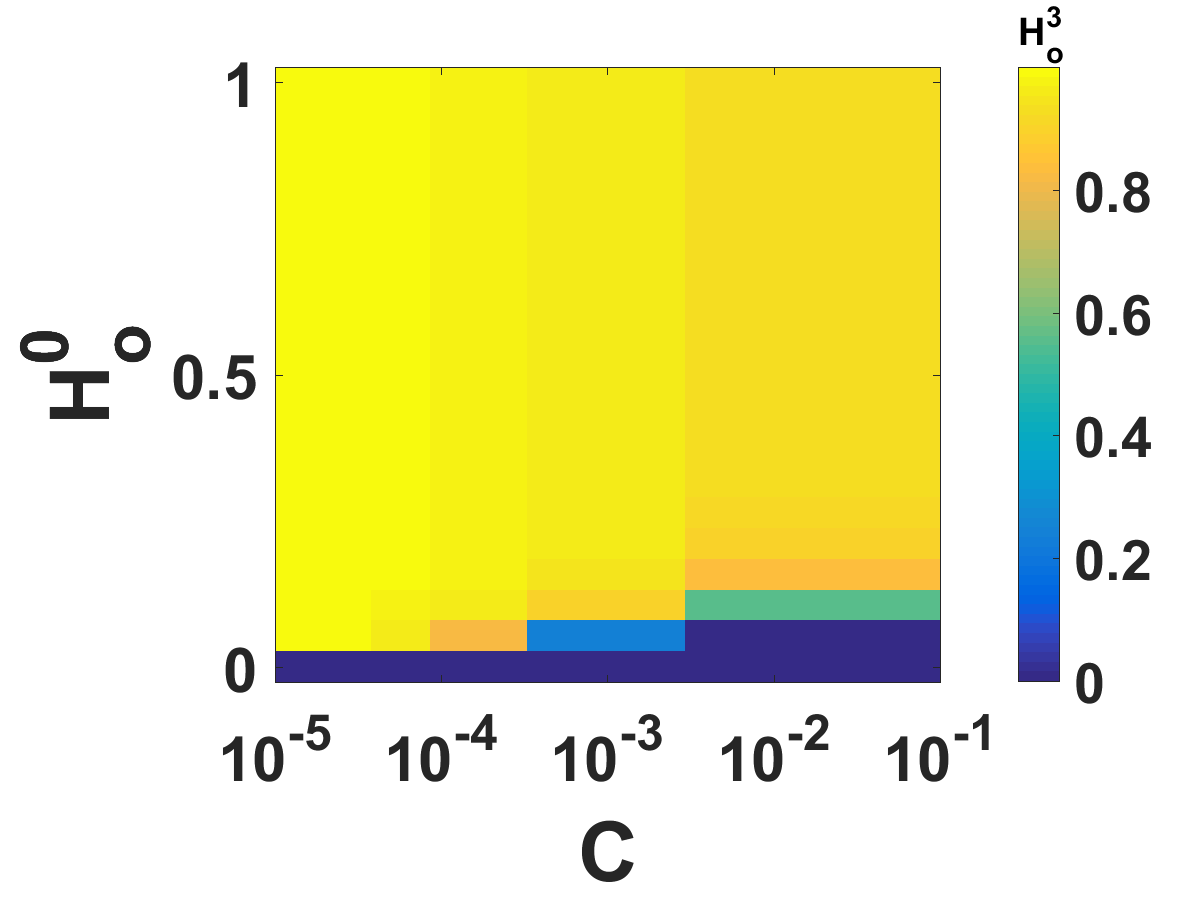} &
\includegraphics[width=40mm]{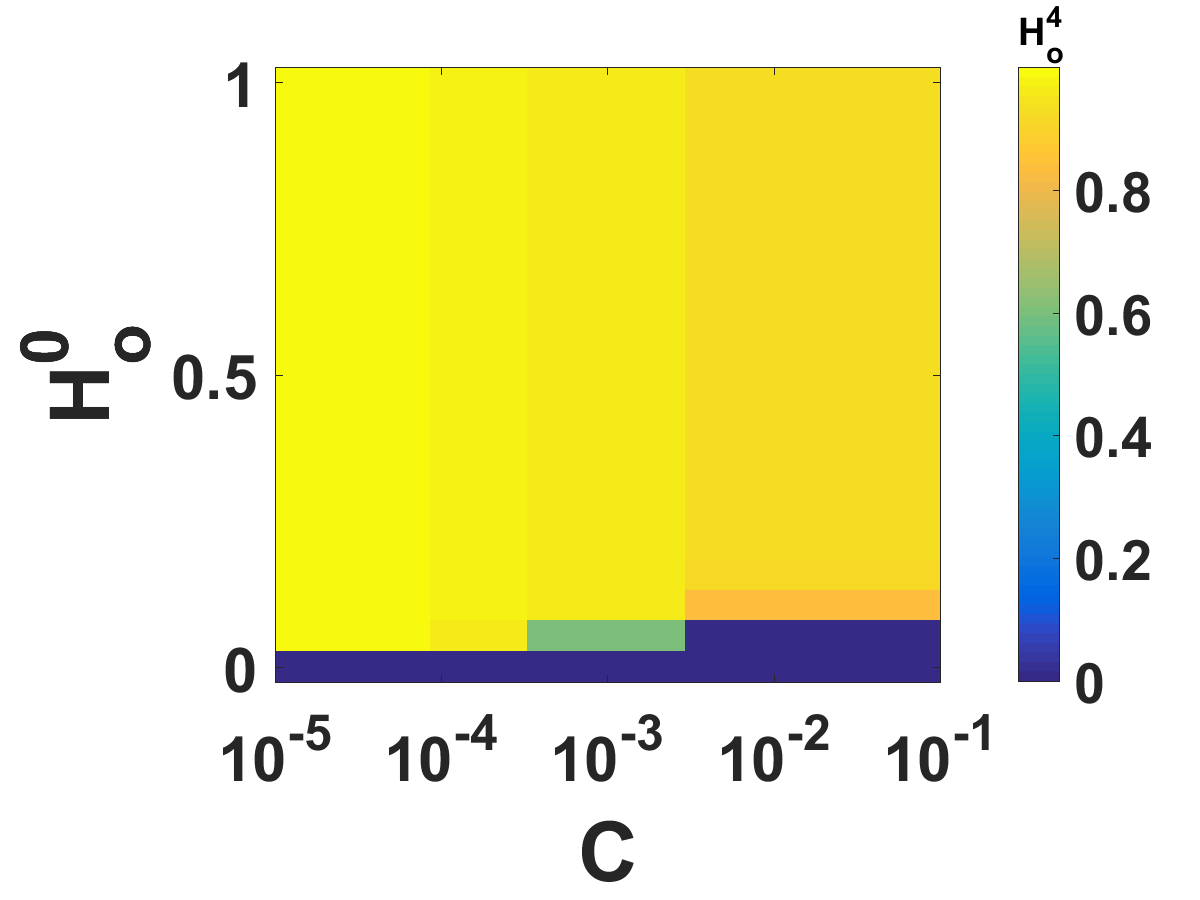} &
\includegraphics[width=40mm]{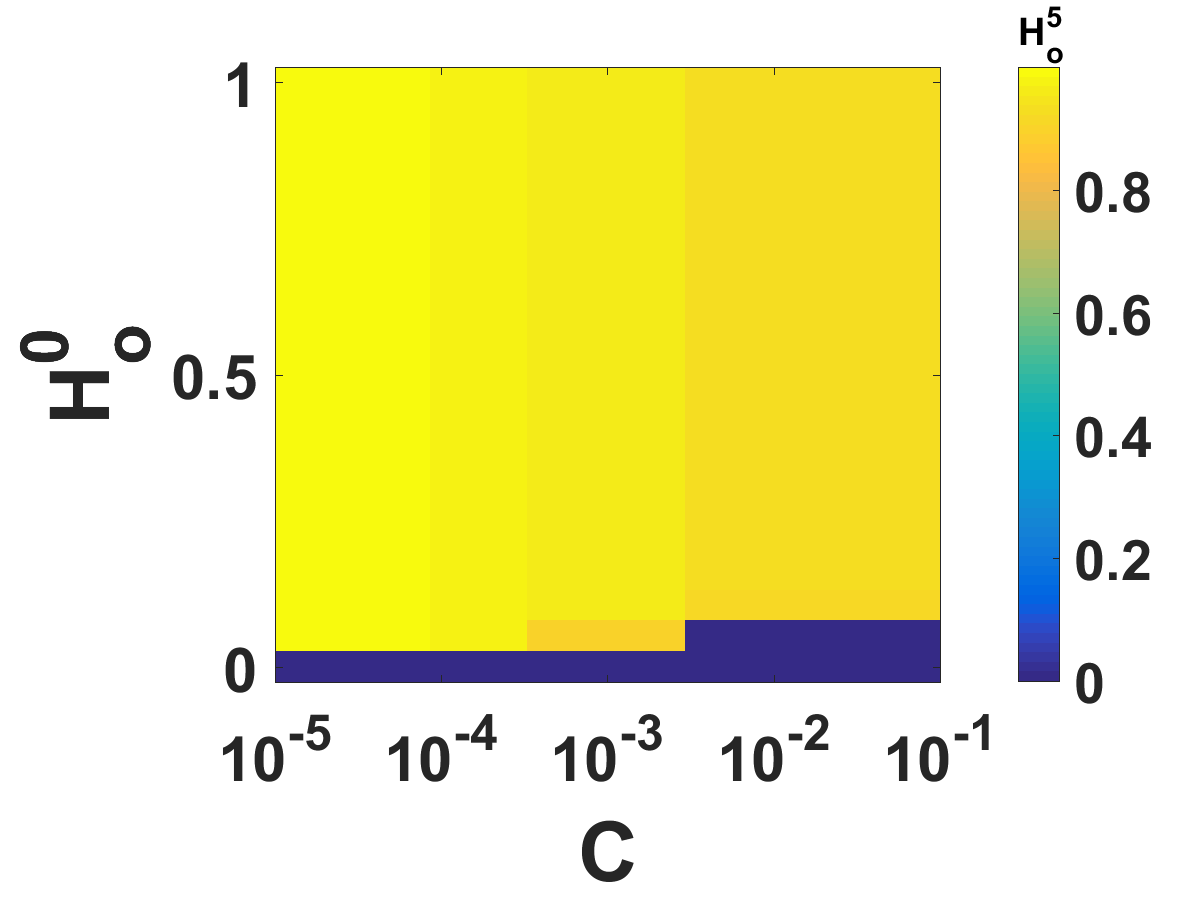} \\
Iteration 3 & Iteration 4 & Iteration 5
\end{tabular}
\caption{Visualization of the convergence of the oracle kernel coefficients to their final values $0$ or $1-C$, and its dependence on the regularization coefficient $C$. Each PROBE iteration ($l = 0 \dots 5$) is depicted as a map, where the coordinates are
($C$, $H_o^0$), and the corresponding map color is $H_o^l$. Kernel coefficient $H_o^l$ values iteratively cluster towards the two stable solutions $0$ or $1-C$. Convergence grows faster as $C$ is decreased.}
\label{fig:psf_convergence_vs_C}
\end{figure}

The rate in which the oracle kernel coefficients $H_o^l$ iteratively converge to their stable solutions $0$ or $1-C$ varies with the regularization parameter $C$. Figure~\ref{fig:psf_convergence_vs_C} displays six PROBE iterations ($l = 0 \dots 5$) as a map, where the map coordinates are ($C$, $H_o^0$), and the corresponding map color is $H_o^l$. At iteration zero $H_o^{l=0} = H_o^{0}$, meaning that for any initial kernel coefficient $H_o^{0}$ value ({\em i.e.,} any row), all the current kernel coefficients $H_o^{l=0}$ values are obviously the same for all $C$. After the first iteration the kernel coefficient $H_o^{l=1}$ values begin  changing towards the stable solutions $0$ or $1-C$; for very small $C$ all of the coefficient $H_o^{l=1}$ values have already converged to either $0$ or $1-C$. Observe that for larger $C$ in iteration $l=1$, the kernel coefficient $H_o^{l=1}$ values have not yet converged to their final bi-level values. In iteration $l=5$ most of the kernel coefficient $H_o^{l=5}$ values have converged to either $1-C$ or $0$. Overall, the rate of $H_o^l$ convergence, and therefore PROBE's convergence rate, increase as we decrease $C$.

The next section analyzes the dynamics of PROBE's residual blur removal process, elucidating additional aspects of its convergence.

\section{PROBE Residual Error Dynamics} \label{sec:PROBE_mse}

In PROBE, we refer to the difference between the restoration outcome in the current iteration and true latent image  as {\em residual} error. The residual error evolves along the iterative process, ideally decreasing until convergence. Following iteration $l$, combining the oracle model~(\ref{eq:oracle_problem}) with the non-blind deblurring process,  the residual blur can be expressed as
\begin{equation}
\epsilon^l = u - \hat{u}^l = u \ast (\delta - h_o^l \ast h_{\mbox{\tiny{RI}}}^l) - n^l \ast h_{\mbox{\tiny{RI}}}^l ~.
\end{equation}
Viewing $u$, $n^l$ as random processes, the current mean squared residual error is
\begin{equation}
\mbox{MSE}^l \triangleq \mbox{MSE}(\epsilon^l) = R_{\epsilon} (\mathbf{0}) = \int
\left[S_u|1 - H_o^l H_{\mbox{\tiny{RI}}}^l|^2 + S_n^l|H_{\mbox{\tiny{RI}}}^l|^2\right]d\mathbf{f} ~,
\end{equation}
where $R_{\epsilon}(\boldsymbol{\tau})$ is the residual error autocorrelation, and $S_u$ and $S_n^l$ are the power spectral densities (PSD) of the latent image $u$ and noise $n^l$ respectively.

Substituting $H_{\mbox{\tiny{RI}}}$ at the respective iteration, the MSE at the current and next iterations can be expressed as
\begin{equation}\label{eq:mse_main}
\mbox{MSE}^l = \int \left[
\frac{S_uC^2}{(|H_o^l|^2 + C)^2} +  \beta^l S_n^l
\right]d\mathbf{f}
\end{equation}
and
\begin{equation} \label{eq:mse_lplus1}
\mbox{MSE}^{l+1} = \int
\left[\alpha^l \frac{S_u C^2}{(|H_o^l|^2 + C)^2} +
\beta^{l+1} \beta^l S_n^l \right]d\mathbf{f}
\end{equation}
\begin{comment}
where
$\alpha\Iter{l} = \frac{(|H_o\Iter{l}|^{2} + C)^2}{(|H_o\Iter{l+1}|^{2} + C)^2}$
and
$\beta\Iter{l} = \frac{|H_o\Iter{l}|^{2}}{(|H_o\Iter{l}|^2 + C)^2}$ (derivations in Appendix~\ref{sec:ab_convergence}).\\
\end{comment}
where
\begin{equation} \label{eq:alpha_beta}
\alpha^l = \frac{(|H_o^l|^2 + C)^2}{(|H_o^{l+1}|^2 + C)^2} \quad \textrm{and} \quad
\beta^l = \frac{|H_o^l|^2}{(|H_o^l|^2 + C)^2} ~~~.
\end{equation}

We define PROBE's boost factor $\mathcal{B}^l \triangleq MSE^l - MSE^{l+1}$
is the measure of error reduced between consecutive PROBE iterations.
We wish to ensure $\mathcal{B}^l > 0$, implying reduction of the residual error between consecutive iterations (boosting). Substituting
equations~(\ref{eq:mse_main}) and~(\ref{eq:mse_lplus1}) we obtain
\begin{equation} \label{eq:MSE_diff}
\mathcal{B}^l =
\int
S_u \cdot (1 - \alpha^l)\frac{C^2}{(|H_o^l|^2 + C)^2}d\mathbf{f} -
\int
S_n^l \cdot (\beta^{l+1} - 1)\beta^l d\mathbf{f}  ~,
\end{equation}
where the first (left) integral is a signal term and the second (right) integral is a noise term.
Since equation~(\ref{eq:MSE_diff}) expresses $\mathcal{B}^l$
as the difference between the signal and noise term,
we wish to increase the signal term and reduce the noise term.

Consider the noise term. Since the noise power $S_n^l$ is multiplied by ${(\beta^{l+1}-1)\beta^l}$, and since $\beta^l$ is non-negative, the noise term contributes to reduction of the residual error when  $\beta^{l+1}<1$.
From equations~~(\ref{eq:oracle_recursive}), (\ref{eq:H_conv})~and ~(\ref{eq:alpha_beta}),
we derive conditions on $C$ and $H_o^{0}$ in order to ensure $\beta^l<1$:
%Combining this condition with the definition of $\beta^l$ (\ref{eq:alpha_beta}) and noting that $|H_o^l|$ is non-negative and $C
\begin{equation}\label{eq:beta_condition}
H_o^{0}\in (0,C)\vee (1-C,1)\,\,\rightarrow\,\,H_o^{l+1} < H_o^l\,\,\rightarrow\,\,
\frac{|H_o^{l}|}{|H_o^{l}|^2 + C} < 1\,\,\rightarrow\,\,\beta^l<1
\end{equation}
In the signal term,
since the signal power $S_u$ is multiplied by $(1 - \alpha^l)$, the signal term is positive when $\alpha^l < 1$, contributing to residual noise reduction.
From equations~(\ref{eq:H_conv}) and~(\ref{eq:alpha_beta}), we derive
conditions on $C$ and $H_o^{0}$  to ensure $\alpha^l<1$:
\begin{equation}\label{eq:alpha_condition}
H_o^{0}\in (C,1-C)\,\,\rightarrow\,\,H_o^{l+1} > H_o^l\,\,\rightarrow\,\,\alpha^l<1
%C<H_o<1-C.
\end{equation}

Taken together, the constraints~(\ref{eq:beta_condition}) and~(\ref{eq:alpha_condition}) on $C$
are sufficient but not necessary. In fact, they are mutually exclusive. At high SNR cases, where
$S_u >> S_n$, we prefer to increase the (large) signal term (by decreasing $C$) even
if it increases the (small) noise term.  However, the noise term
increases with PROBE iterations motivating an higher $C$. Nevertheless in this work our analysis assumes a constant $C$ throughout the iterative PROBE process.

Once convergence has been reached,  $H^{l+1} = H^l$. Substituting in~(\ref{eq:alpha_beta}),
\begin{equation}
\lim_{l \rightarrow \infty} \alpha^l = 1 \quad \textrm{and} \quad \lim_{l \rightarrow \infty} \beta^l = 1 \\
\end{equation}
Thus, after PROBE convergence, once $\alpha$ and $\beta$ equal $1$, both the signal and noise terms fall out, leaving
$\mathcal{B}^\infty = 0$, {\em i.e.,} further boosting is not possible.

We have now derived conditions for residual error reduction in PROBE, and shown
that convergence corresponds to stopping when the residual error cannot be further reduced.
Recall, however, that the analysis in sections~\ref{sec:PROBE_analysis} and~\ref{sec:PROBE_mse} relies on modelling the blur-kernel estimation model by an oracle. The next section addresses the applicability of the model-based analysis to a real PROBE system.

\section{From Theory to Practice} \label{sec:analysis_simulations}
An actual PROBE system applies a feasible blur-kernel estimator, and the oracle model no longer holds.
In this respect, our PROBE system follows~\cite{hanocka2015} and employs an ad-hoc MAP$_{u_c,h}$ PSF estimation algorithm extracted from the implementation of Kotera {\em et al}~\cite{kotera2013}.
Compared to the oracle, the MAP$_{u_C,h}$ PSF estimator adds an extra layer of error, appearing in the PROBE iterations as
\begin{equation}\label{eq:ad_hoc_error}
g^l =
(h_{\textrm{MAP}_{u_c,h}}^l + n_{\textrm{MAP}_{u_c,h}}^l)\ast u + n^l =
h_{\textrm{MAP}_{u_c,h}}^l\ast u + \tilde{n}^l,
\end{equation}
where $\tilde{n}^l \triangleq  n_{\textrm{MAP}_{u_c,h}}^l\ast u + n^l$.
Therefore, using an ad-hoc kernel estimation $h_{\textrm{MAP}_{u_c,h}}$ rather than an oracle leads to higher effective noise
$\tilde{n}$.

In our experiments we usually report PSNR (peak signal-to-noise ratio) instead of MSE, since it is more commonly used in image quality assessment (using the tools from~\cite{Kohler2012}). PROBE residual MSE is readily converted to PSNR
\begin{equation}\label{eq:psnr}
\text{PSNR}^l = -10\log_{10}\text{MSE}^l,
\end{equation}
since all gray levels are scaled to the range $[0,1]$.

To corroborate our analysis, we compare the predicted, simulated and experimental PSNR obtained along PROBE iterations.
\begin{itemize}
\item
Given a test image with synthetic blur and known additive noise, and assuming oracle PSF estimation in PROBE,
we predicted $\mbox{MSE}^l$ (hence $\mbox{PSNR}^l$) using equation~(\ref{eq:mse_main}), where we estimated $S_u$  from the image, and evolved $S_n^l$ from the spectral density $S_n$ of the additive noise.
\item
With the same blurred and noisy image, again assuming the oracle model, we  simulated PROBE using
equation~(\ref{eq:deblur_pseudo_iters}), computing $\mbox{PSNR}^l$ by comparison to the latent image.
\item
We obtained $\mbox{PSNR}^l$ experimentally using an actual PROBE system with
an ad-hoc MAP$_{u_c,h}$ kernel estimator.
\end{itemize}

\begin{figure}[t]
\begin{tabular}{ccc}
Prediction & Simulation & Experiment \\
\includegraphics[width=40mm]{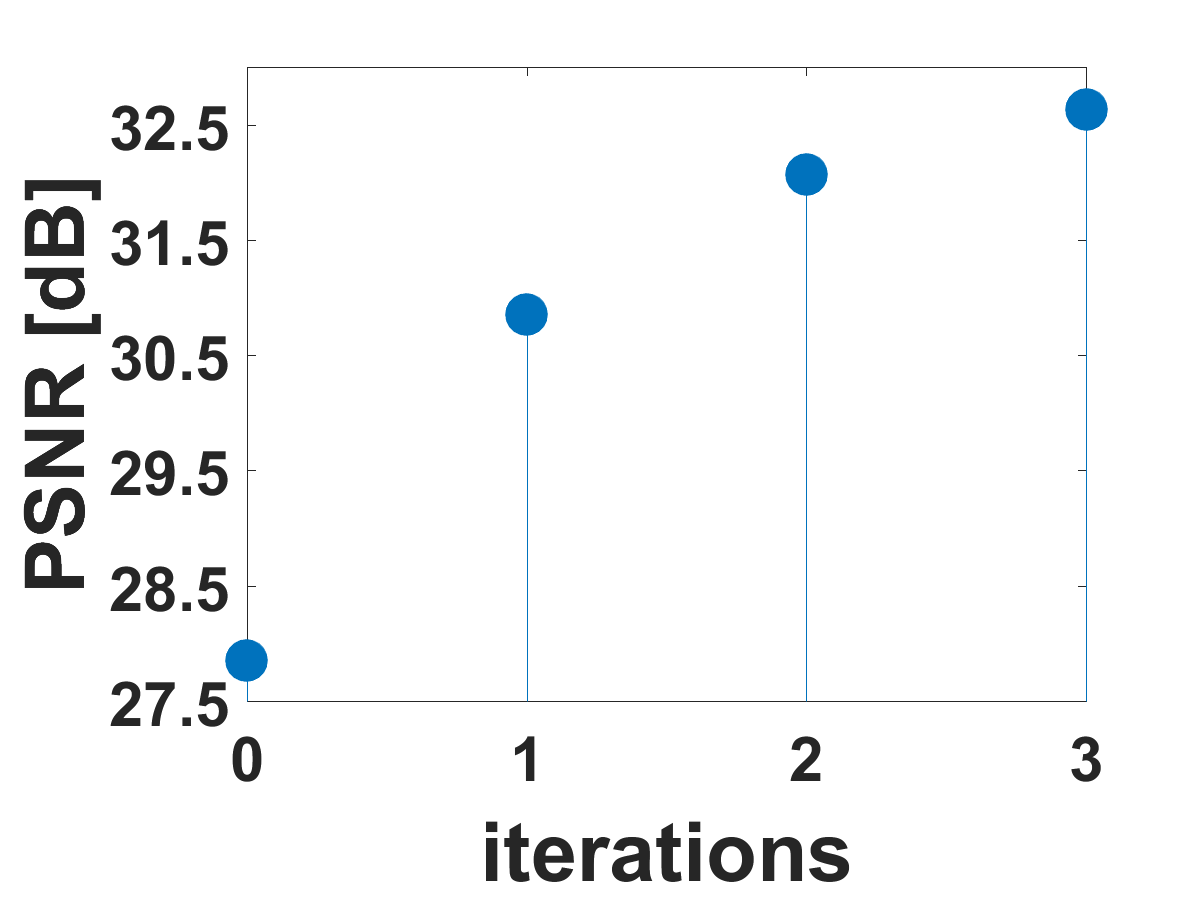} &
\includegraphics[width=40mm]{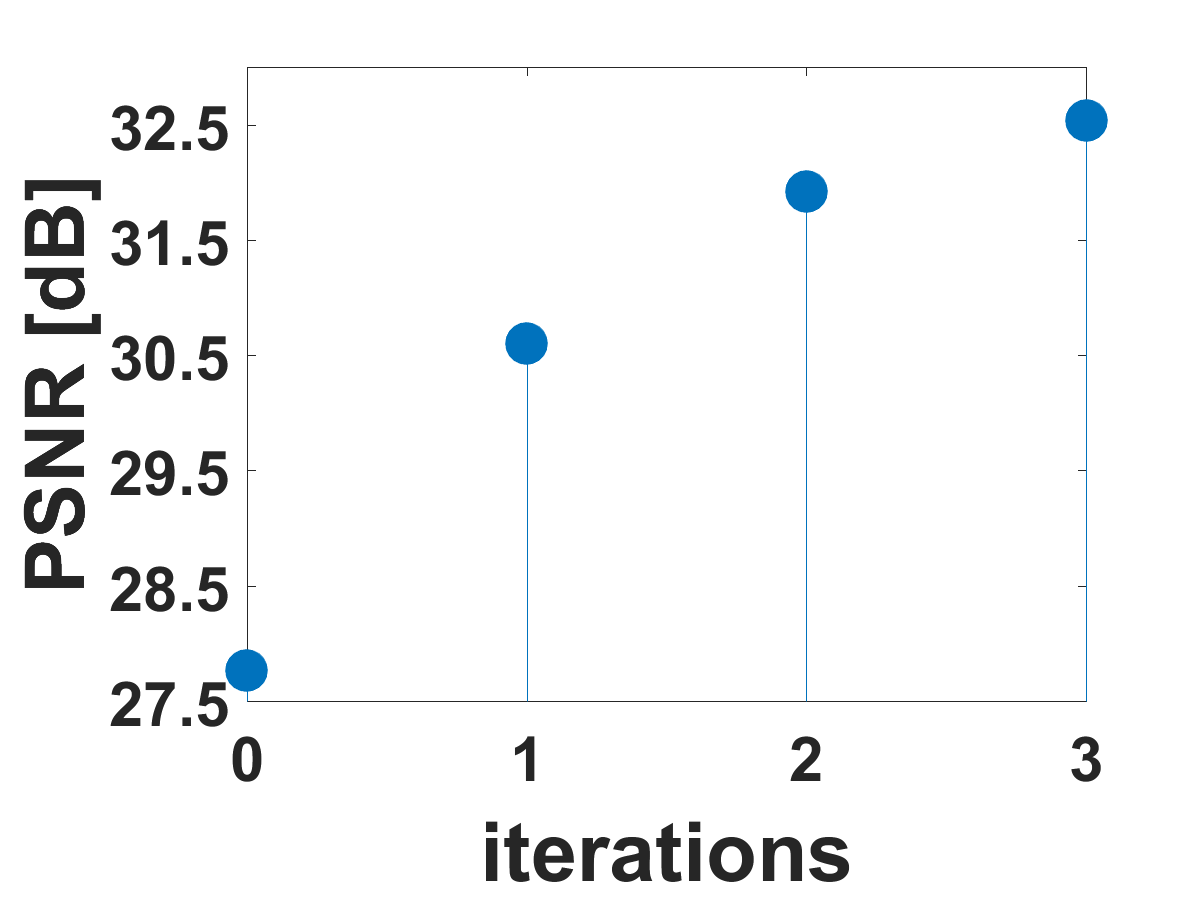} &
\includegraphics[width=40mm]{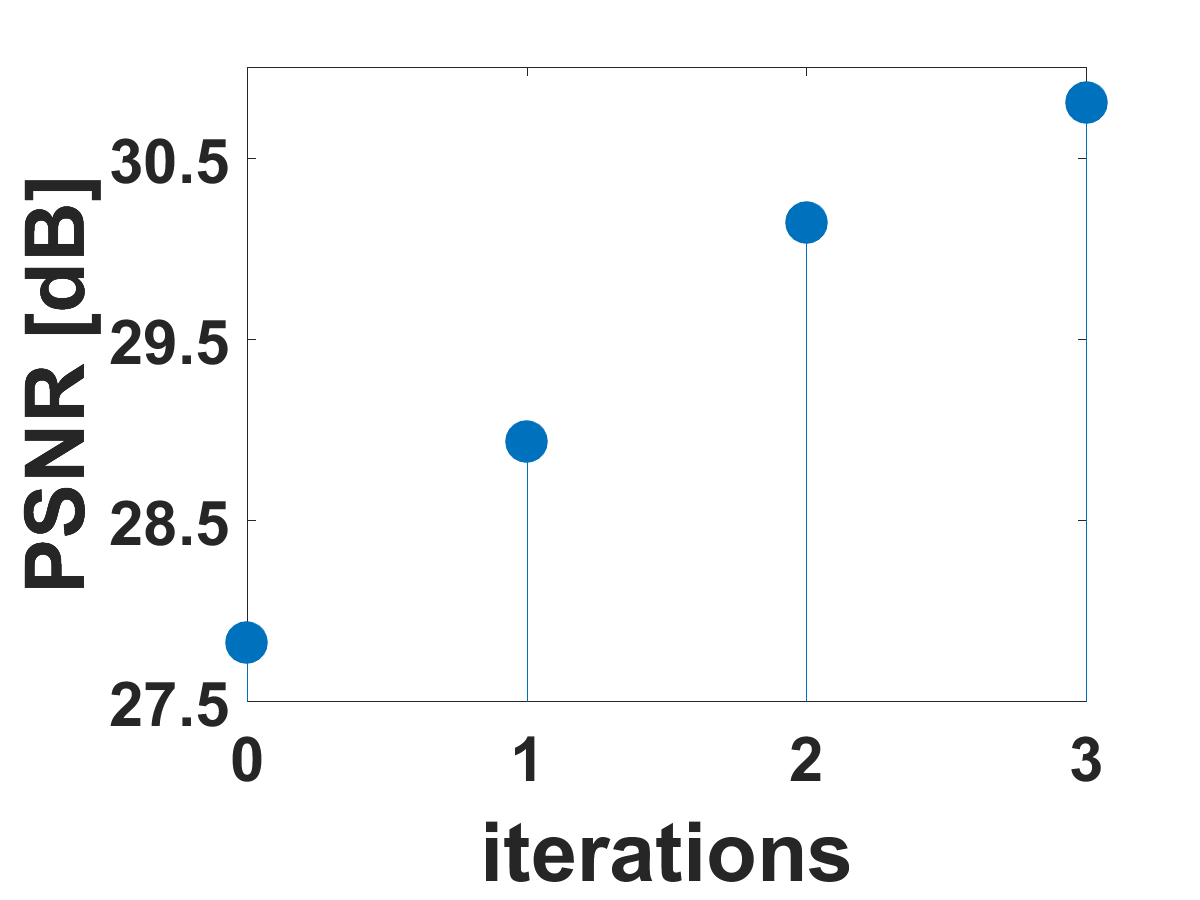}\\
\end{tabular}
\caption{PSNR for the first three PROBE iterations, based on analytic prediction (left), simulation (middle) and experimental results (right). Successful elimination of residual blur, and PSNR improvement, are consistently observed. The prediction and simulation results, based on the oracle model, are slighty superior to the experimental results. }
\label{fig:psnr_compare}
\end{figure}

The predicted, simulated and experimental results of three PROBE iterations are compared in
Fig.~\ref{fig:psnr_compare}. In all three evaluation approaches, PSNR improves in the iterative process.
The prediction and simulation results, based on the oracle model, are very similar. They are both about 2dB better than the experimental results, obtained using a real PSF estimator rather than the oracle model.

\begin{figure}[t]
\begin{center}
\begin{tabular}{cc}
Analytical & Experiments \\
\includegraphics[width=55mm]{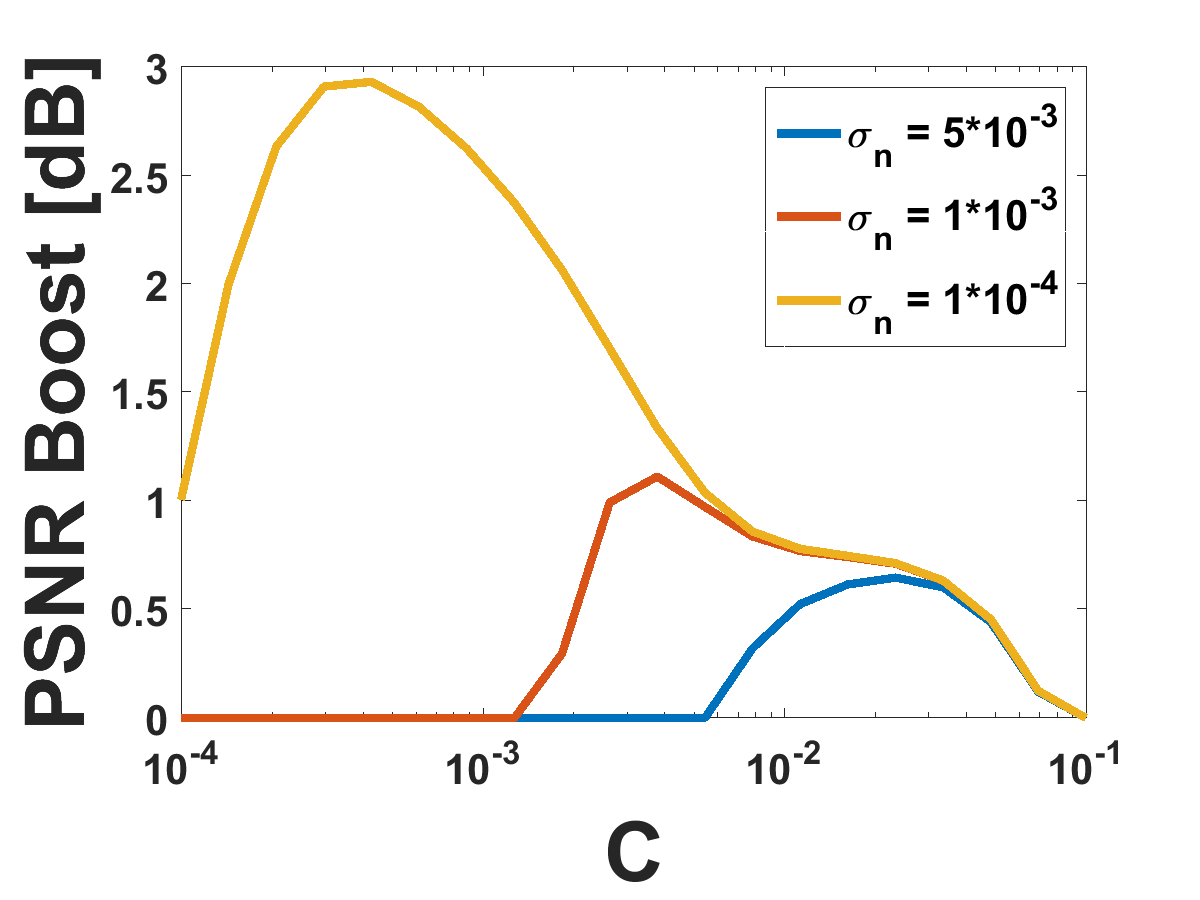} &
\includegraphics[width=55mm]{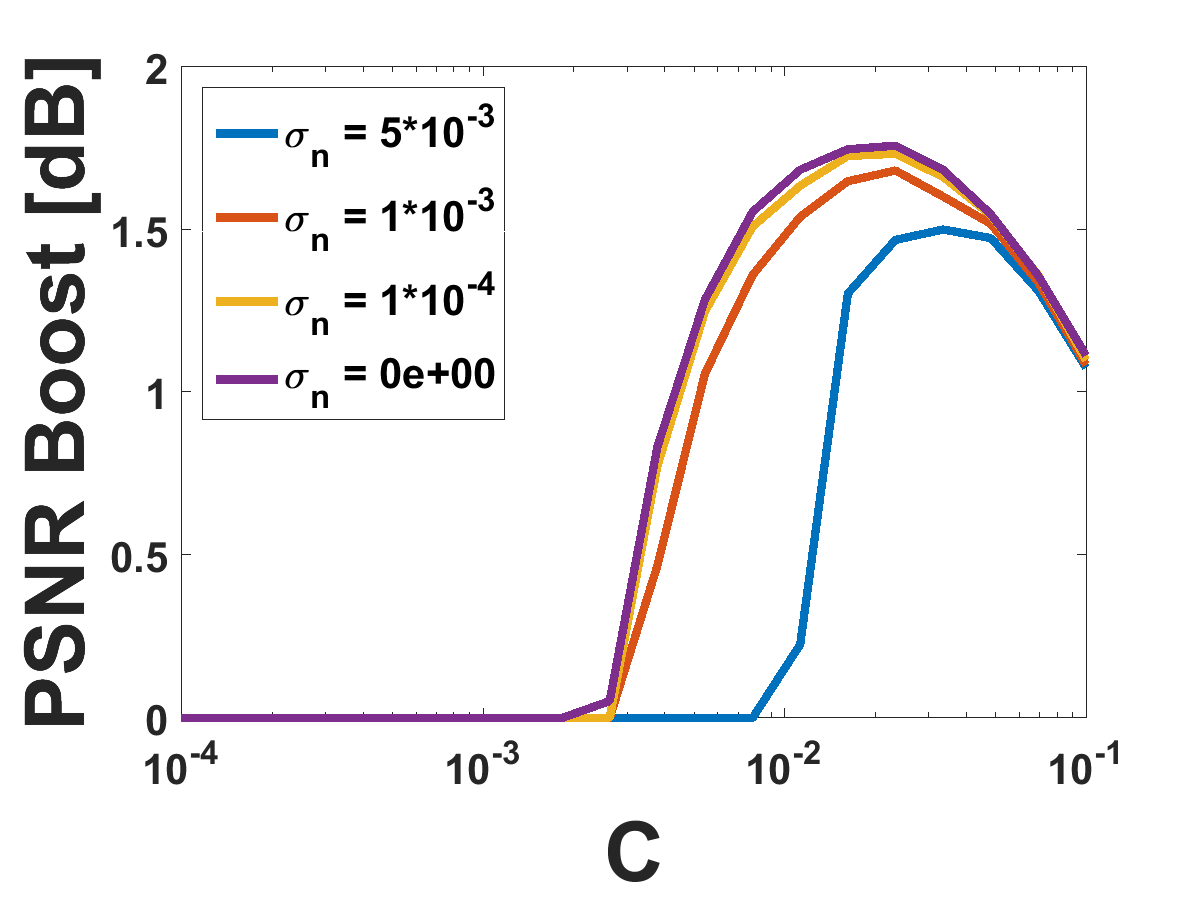} \\
\end{tabular}
\end{center}
\caption{PSNR boost in the 2nd PROBE iteration, breaking the limit of the sequential approach.
Noise ($\sigma_n$) effects the PSNR boost and the range of possible $C$ values. The real experiment is subject
to an effective noise floor, due to imperfect kernel estimation.}
\label{fig:psnr_improvement_noise}
\end{figure}

A sequential algorithm for non-blind deblurring, consisting of blur-kernel estimation followed by non-blind deblurring, typically improves PSNR. The sequential approach coincides with the first PROBE iteration. With PROBE, additional iterations further boost the PSNR. The actual boost depends on the level of additive noise.
Fig.~\ref{fig:psnr_improvement_noise}(left) shows the analytically predicted PSNR boost, using the oracle blur-kernel estimation model, as a function of $C$, for several values of $\sigma_n$, where $S_n^0 = \sigma_n^2$.
Fig.~\ref{fig:psnr_improvement_noise}(right) shows the corresponding graphs for an actual experimental setup.
The imperfect PSF estimation in the experimental case creates an effective noise floor, as described by equation~(\ref{eq:ad_hoc_error}), and that noise poses an effective lower limit on useful values of the regularization parameter $C$.

\section{Experimental Evaluation}

\begin{figure}[t]
\makebox[\textwidth][c]{
\begin{tabular}{cccc}
Blurred Image & Iteration 1 & Iteration 2 & Iteration 3 \\
\includegraphics[width=0.25\textwidth]{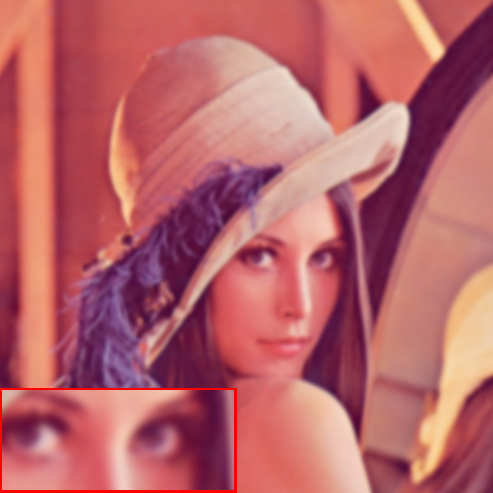}  &
\includegraphics[width=0.25\textwidth]{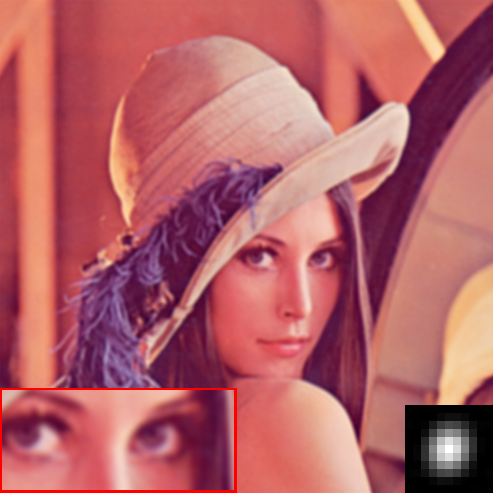} &
\includegraphics[width=0.25\textwidth]{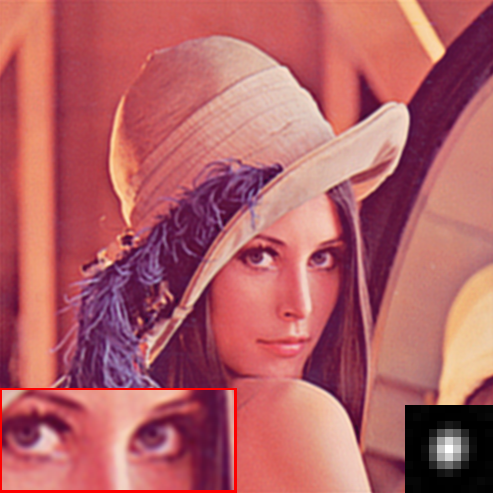} &
\includegraphics[width=0.25\textwidth]{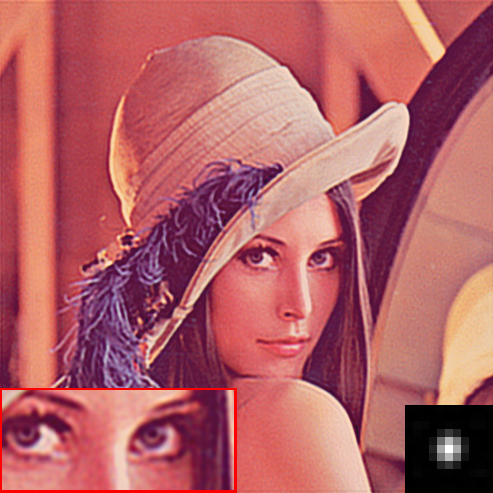} \\
\end{tabular}
}%
\caption{PROBE deblurs Lena. Displayed: the blurred image and PROBE results by iteration with corresponding residual blur kernels superimposed.}
\label{fig:feat_ex1}
\end{figure}

\begin{figure}[t]
\makebox[\textwidth][c]{
\begin{tabular}{cccc}
Blurred Image & Iteration 1 & Iteration 2 & Iteration 3 \\
\includegraphics[width=0.25\textwidth]{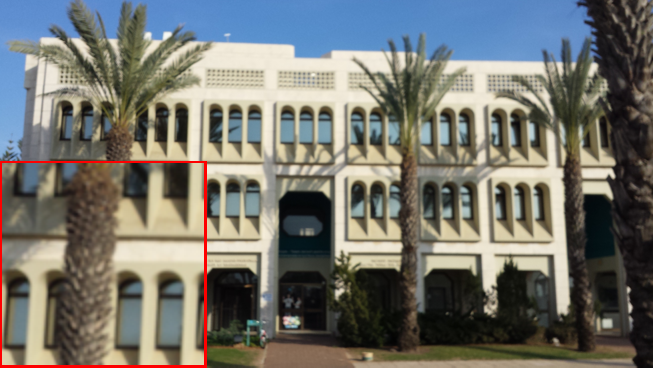}  &
\includegraphics[width=0.25\textwidth]{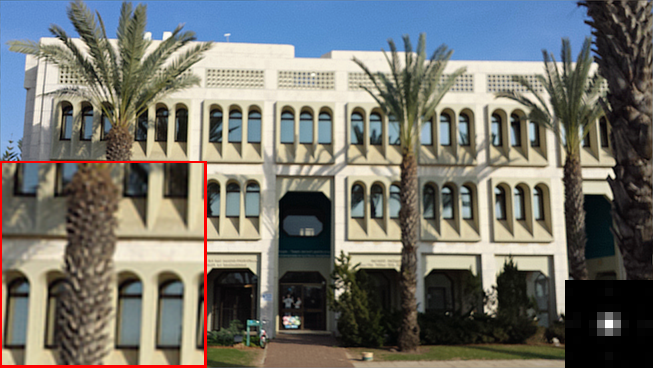} &
\includegraphics[width=0.25\textwidth]{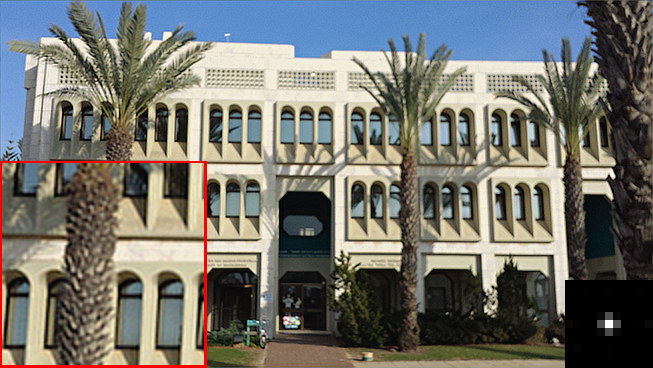} &
\includegraphics[width=0.25\textwidth]{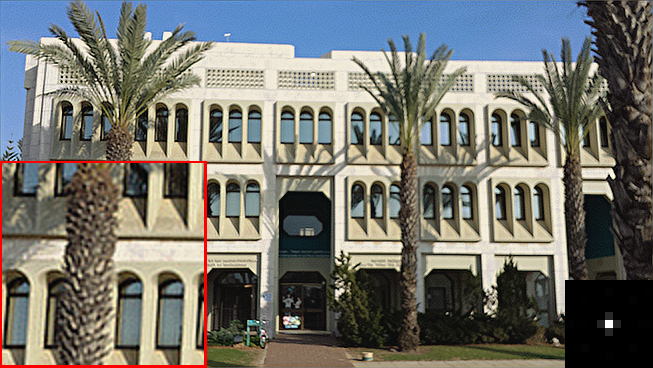} \\
\end{tabular}
}%
\caption{PROBE deblurs an out-of-focus image obtained with a smartphone. Displayed: The blurred image and PROBE results by iteration with the corresponding shrinking residual blur kernels superimposed.}
\label{fig:feat_ex2}
\end{figure}

PROBE's building-blocks are an ad-hoc MAP$_{u_C,h}$ kernel estimator extracted from the implementation of Kotera {\em et al}~\cite{kotera2013} and the low cost modified inverse filter~(equation~\ref{eq:pseudo-inverse}). Typical PROBE operation is shown in Figs.~\ref{fig:feat_ex1} (blurred Lena) and~\ref{fig:feat_ex2} (out-of-focus image acquired using a smartphone). We show the visual improvement by iteration, and the shrinkage of the residual blur kernels.

%Featured Example 1: Intermediate PROBE - TAU

\begin{figure*}[t]
\makebox[\textwidth][c]{
\begin{tabular}{cc}
\includegraphics[width=0.37\textwidth]{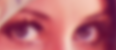} &
\includegraphics[width=0.37\textwidth]{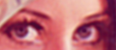} \\
Blurred Image (PSNR: 28.09) & PROBE (PSNR: \textbf{32.32})\\
%PSNR: 28.09 & PSNR: \textbf{32.32} \\
\includegraphics[width=0.37\textwidth]{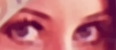} &
%\includegraphics[width=0.3\textwidth]{kotera_kotera_lena_color_512_deblur_itr1} \\
%Krishnan {\em et al}~\cite{krishnan2011} (PSNR: 30.54) & Kotera {\em et al}~\cite{kotera2013} (PSNR: 30.51 ) \\

\includegraphics[width=0.37\textwidth]{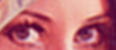} \\
Krishnan~\cite{krishnan2011} (PSNR: 30.54) & Hanocka~\cite{hanocka2015} (PSNR: 30.60 ) \\

\includegraphics[width=0.37\textwidth]{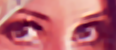} &
\includegraphics[width=0.37\textwidth]{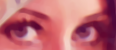} \\
Shan~\cite{shan2008} (PSNR: 29.20) & Kotera~\cite{kotera2013} (PSNR: 30.51 ) \\
\end{tabular}
}%
\caption{Gaussian blurred Lena and the restoration results obtained using PROBE and four state-of-the-art blind deblurring algorithms. Only the eye region is shown. Note the pupils.}
\label{fig:showcase_Lena_vs_others}
\end{figure*}

Figure~\ref{fig:showcase_Lena_vs_others} compares PROBE's deblurring result to  sophisticated state of the art
methods~\cite{krishnan2011,kotera2013,shan2008,hanocka2015}. Starting with the same Gaussian blurred Lena image,  PROBE, using a modified inverse filter as its non-blind deblurring module, yields the best visual outcome (compare pupils)
and the highest PSNR.

%SHOWCASE: Lena VS state-of-the-art

%SHOWCASE: Compare Levin Results
\begin{figure*}[t]
\makebox[\textwidth][c]{
\begin{tabular}{ccc}
\includegraphics[width=0.3\textwidth]{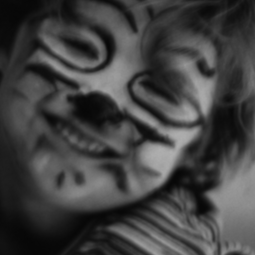} &
\includegraphics[width=0.3\textwidth]{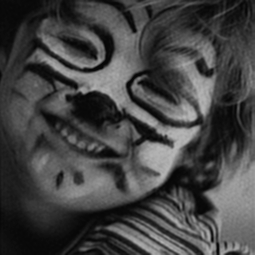} &
\includegraphics[width=0.3\textwidth]{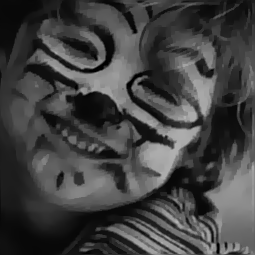} \\
Blurred Image & PROBE (PSNR: \textbf{34.74}) & Cho~\cite{SCho_deblur_2009} (PSNR: 30.81) \\

\includegraphics[width=0.3\textwidth]{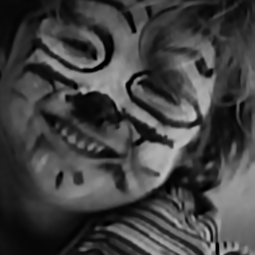} &
\includegraphics[width=0.3\textwidth]{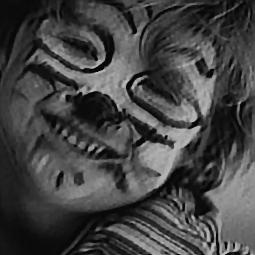} &
\includegraphics[width=0.3\textwidth]{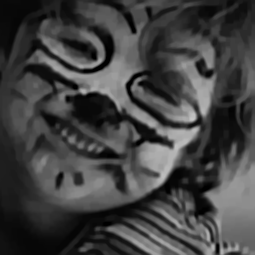} \\
Levin~\cite{levin2011} (PSNR: 31.73) & Shan~\cite{shan2008} (PSNR: 25.26) & Krishnan~\cite{krishnan2011} (PSNR: 30.67)\\
\end{tabular}
}
\caption{Featured example from dataset~\cite{levin2009}}
\label{fig:showcase_Levin_vs_others}
\end{figure*}

We provide systematic quantitative performance evaluation using the rigorous dataset of
Levin {\em et al}~\cite{levin2009}. Fig.~\ref{fig:showcase_Levin_vs_others} shows one blurred image from the dataset and its restoration using PROBE and alternative state of the art methods.

For quantitative comparison across the dataset we employ the error ratio metric, defined as
$\frac{\textrm{SSD}(u,\hat{u})}{\textrm{SSD}(u, \hat{u}|h)}$ where the sum of squared differences (SSD) is equal to MSE up to a scalar factor. The numerator is the difference between the latent image and its blind restoration using the method under test. The denominator is the difference between the latent image and a reference non-blind restoration result. It is customary to present the error ratio statistic over the whole dataset as the cumulative distribution of individual image error ratios.

A caveat in the error ratio measure is the non-blind reference algorithm used to compute the denominator. In straightforward sequential algorithms, it makes sense to use the non-blind deblurring module of the respective blind algorithm. This leads to a {\em relative} error ratio measure, that highlights the added value of the blind deblurring scheme with respect to its non-blind component. However, several recent
publications~\cite{levin2009,zhou2014,krishnan2011} compare blind deblurring methods using the same non-blind deblurring algorithm in the denominator. In this case, we refer to the outcome as {\em absolute} error ratio.

\begin{figure}[t]
\begin{tabular}{cc}
\includegraphics[width=60mm]{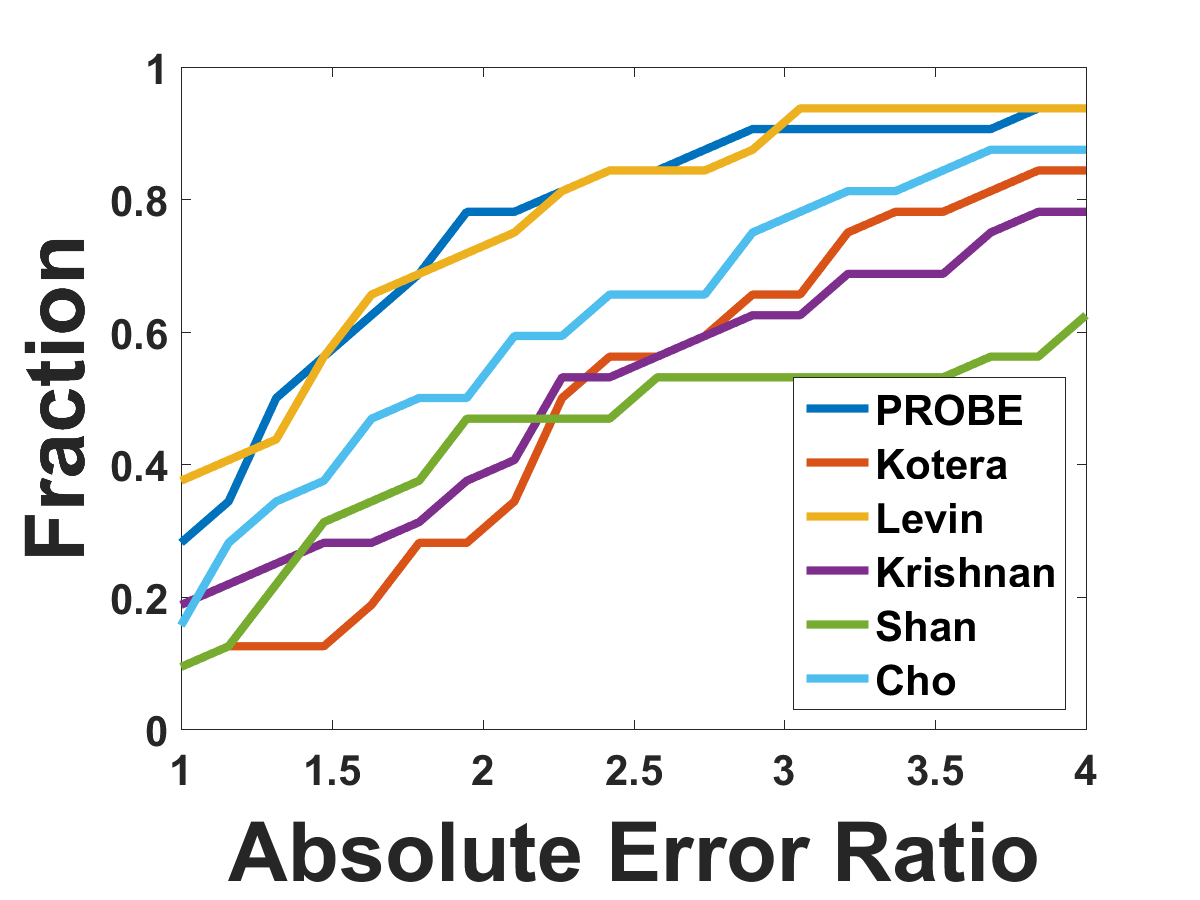} &
\includegraphics[width=60mm]{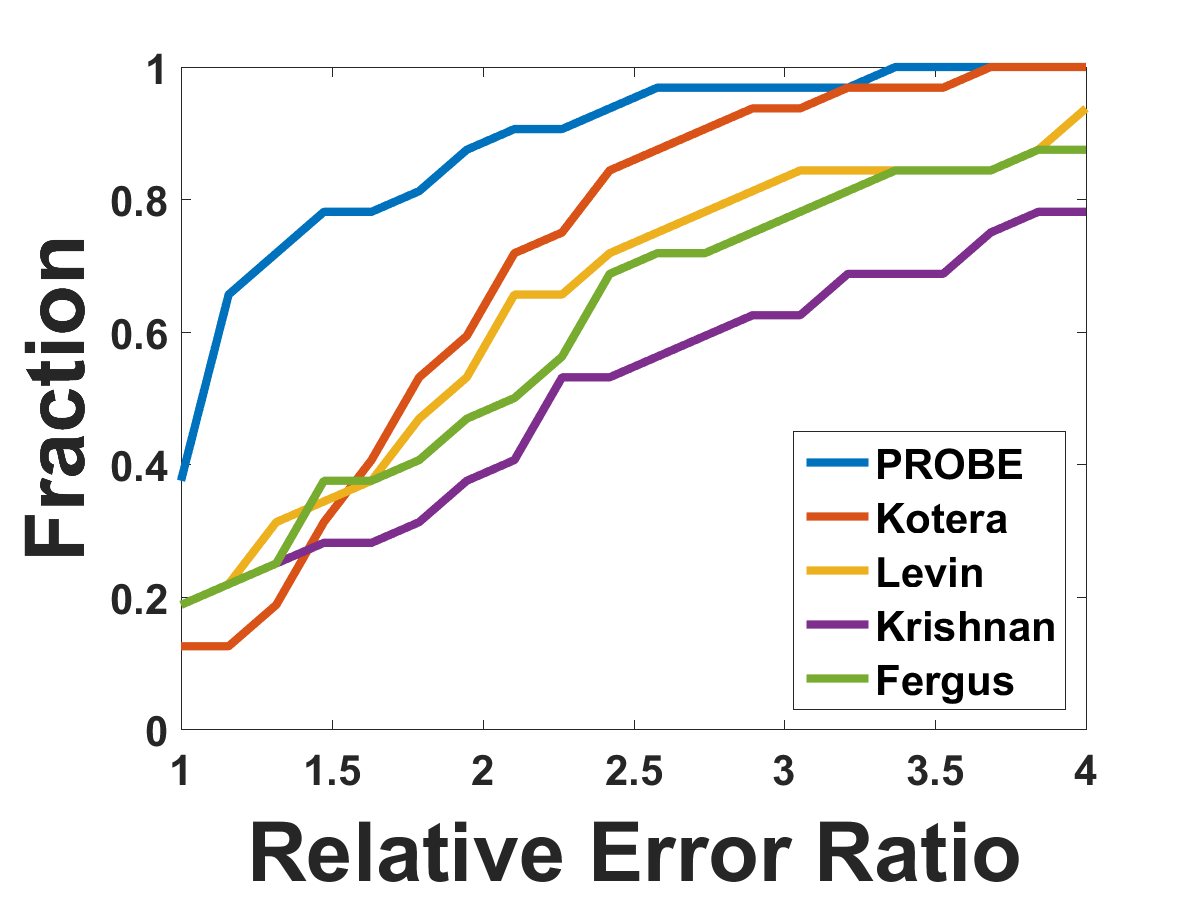} \\
\end{tabular}
\caption{Absolute Error ratio and Relative Error ratio on dataset~\cite{levin2009}}
\label{fig:errorRatio}
\end{figure}

Fig.~\ref{fig:errorRatio}(left and right) show the absolute and relative error ratio of PROBE and state of the art algorithms respectively. Krishnan {\em et al}~\cite{krishnan2009} is used as the reference non-blind deblurring algorithm for absolute error ratio calculation. Since PROBE relies on the simple modified inverse filter, its relative error ratio is vastly superior to other methods, that use sophisticated non-blind deblurring components. PROBE's absolute error ratio is similar to~\cite{levin2011}, both algorithms being superior to other methods.

%For each method we computed mean PSNR on the dataset, we recorded the ranking of each method, from 1 (best) to 5, by PSNR.
%For each image in the dataset, we recorded the ranking of each method, from 1 (best) to 5, by PSNR.
In Table~\ref{fig:ranktable} we present the average PSNR for each method and corresponding MATLAB run time\footnote{Using a single-threaded Matlab on a 3.4Ghz CPU.}.
PROBE's mean PSNR achieves state of the art results. Our unoptimized MATLAB implementation is $7$ times faster than~\cite{levin2011}.
%and a final rank derived from the mean rankings. PSNR differnces between the
%best and second best methods on a given image were often small, sometimes less than 0.1dB.
%Notably, the method of Kotera {\em et al} performed best on no less than 43.8\% of Levin's database~\cite{levin2009},
%yet PROBE had the best mean rank and mean PSNR, indicating consistent overall performance.

\begin{table}[t]
\center
\begin{tabular}{ |l||c|c|}
\hline
 Method & Mean PSNR & Time [sec] \\
\hline
  Kotera~\cite{kotera2013} & 26.3 & 3 \\
  Krishnan~\cite{krishnan2011} & 26.5 & 9 \\
  Shan~\cite{shan2008} & 25.4 & NA \\
  Cho~\cite{SCho_deblur_2009} & 27.1 & NA \\
  Levin~\cite{levin2011} & 28.4 & 36 \\
  {\bf PROBE} & 28.4 & 5 \\
\hline
\end{tabular}
% \csvautotabular{fullrankTable.csv} \\[0.2cm]
\caption{PSNR for PROBE and modern deblurring methods over the
dataset of~\cite{levin2009}. Average computing time per-frame is reported for MATLAB implementations only.}
\label{fig:ranktable}
\end{table}

\section{Discussion}

PROBE is a novel recursive blind deblurring framework, using a feedback architecture
and the simple modified inverse filter as its deblurring engine.
Comparative performance analysis using the challenging database of
Levin {\em et al}~\cite{levin2009} reveals that PROBE is second to none at the time of writing.

Using a novel oracle model for PROBE's PSF estimation module, we provide analytic convergence and error analysis for PROBE and demonstrate their validity and predictive value.
This is an unusual feat in the blind deblurring literature.

PROBE is ideal for resource-limited computation; we are currently implementing PROBE as a cellphone application. PROBE's computational bottleneck is in its PSF estimation module which, following~\cite{hanocka2015}, is still borrowed from an ad-hoc MAP$_{u_c,h}$ algorithm in the implementation of Kotera {\em et al}~\cite{kotera2013}. Recent literature hints that PSF estimation should not be too difficult by
itself~\cite{levin2009,michaeli2013}. Studying the use of simpler PSF estimation module within PROBE is an interesting direction for future research.

\bibliographystyle{splncs}
\bibliography{probe-arXiv}
\end{document}